\documentclass[11pt]{article}

\usepackage[preprint]{acl}

\usepackage{times}
\usepackage{latexsym}

\usepackage[T1]{fontenc}

\usepackage[utf8]{inputenc}

\usepackage{microtype}

\usepackage{inconsolata}

\usepackage{graphicx}

\usepackage{yu-setup}

\title{Recursive Reasoning or Statistical Extrapolation? In-Context Learning in Multi-Agent Interdependent Decision-Making}

\author{
 \textbf{Yu Liu},
 \textbf{Wenwen Li},
 \textbf{Yifan Dou},
 \textbf{Guangnan Ye}
\\
 Fudan University
\\
\texttt{yuliu23@m.fudan.edu.cn, \{liwwen,yfdou,yegn\}@fudan.edu.cn}
}

\begin{document}
\maketitle
\begin{abstract}
In-context learning (ICL) enables large language model (LLM) agents to improve decisions using interaction history, yet it remains unclear whether such improvement reflects refined internal reasoning or mere extrapolation of statistical patterns. To disentangle these mechanisms, we study LLM agents in multi-agent incomplete-information games that require recursive belief reasoning. By constructing a public goods game and manipulating the statistical structure of historical feedback, we evaluate decision quality against a history-independent rational expectations equilibrium (REE) benchmark. Our experiments reveal that when historical statistical patterns are disrupted, the benefits of longer context largely vanish, degrading decision quality to the no-context baseline in a way sharply amplified by stronger strategic interdependence. These results suggest that, in such strategic environments, ICL behavior is more consistent with statistical extrapolation than with strategic reasoning. Our work extends the mechanistic study of ICL to strategic multi-agent settings, introduces REE as a diagnostic tool for distinguishing reasoning from extrapolation, and provides a reusable framework for probing the boundaries of LLM reasoning in recursive belief tasks. 
\end{abstract}

\section{Introduction}

LLM-based agents have shown strong decision-making performance, with in-context learning (ICL) as a core mechanism \citep{brown2020, wei2022}. By placing interaction histories into the context window, models appear able to adjust their behavioral policies in response to feedback \citep{xia2025}. However, a critical question remains unsettled: Does the behavioral improvement exhibited by ICL originate from strategic corrections of the internal decision-making process at inference time \citep{xie2022}, or is it merely pattern matching and extrapolation of statistical regularities present in the context sequences \citep{olsson2022context}? In single-agent or purely cooperative settings, the two mechanisms are often observationally equivalent and therefore hard to disentangle.

\begin{figure}
    \centering
    \includegraphics[width=0.8\columnwidth]{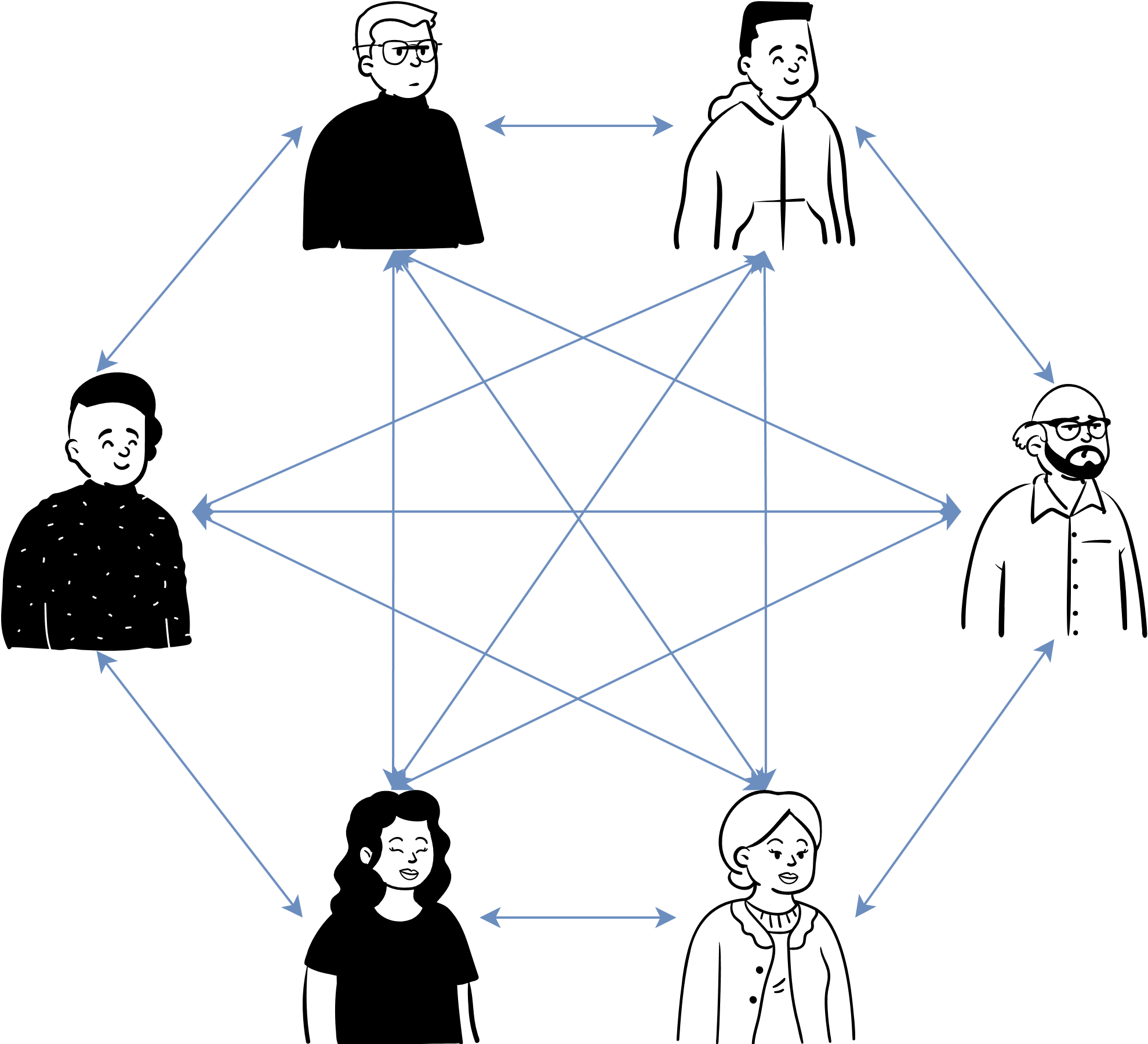}
    \caption{\textbf{Relational Complexity in Multi-Agent Systems.} When everyone is constantly influencing everyone else, making a single decision becomes an overwhelming task. This structure necessitates a shift from simple logic to a deep and complicated cycle of guessing what others might do, highlighting the challenge of strategic coordination.}
    \label{figure:network}
\end{figure}

We break this equivalence by using multi-agent games with strategic interdependence \citep{Harsanyi2004}, where optimal actions depend on expectations about others, requiring recursive belief reasoning. As the degree of interdependence deepens, environmental feedback varies nonlinearly with one another's decisions, and straightforward autoregressive extrapolation along historical trends can lead to systematic biases. This property provides a way to probe the mechanism of ICL.

We concretely construct a repeated $n$-person public goods game \citep{Fehr2000} as an experimental testbed. In each round, after observing a publicly announced cost, every agent independently decides whether to participate in a public project that generates positive externalities. Because REE-derived policies \citep{John1961} depend only on strategic fundamentals and not on historical realizations, they provide a history-independent reference that helps isolate behaviors that are systematically misaligned with recursive reasoning. Adopting the in-context reinforcement learning paradigm, we let multiple populations of LLM agents repeatedly interact in the aforementioned public goods game while systematically manipulating two dimensions: the strength of interdependence $\beta$, and the statistical structure of the environmental feedback sequences. By combining analytical solutions of the theoretical equilibrium with regression analyses of behavioral data, we examine the performance boundaries of ICL under varying demands for recursive reasoning.

The experimental results provide evidence that ICL primarily operates as statistical extrapolation under high interdependence. In scenarios with weak interdependence and simple environmental feedback, it appears to bring about improvements in decision-making behavior. However, the improvement is primarily constrained by the statistical structure of the feedback. Under monotonic trends ICL consistently improves over baseline, while under jump sequences the benefit of longer context largely disappears, with higher $\beta$ further reducing the residual gain. This suggests that, in these multi-agent interdependent scenarios, ICL's behavior appears more consistent with extrapolating along statistical trends than with recursive belief revision under the equilibrium benchmark.

Our contributions are as follows:

\begin{itemize}
    \item We extend the mechanistic study of ICL to multi-agent interdependent games, probe the limits of ICL when recursive belief reasoning is required, and provide testable theoretical constraints for identifying the boundary conditions of recursive belief reasoning in LLMs.
    \item We introduce rational expectations equilibrium into the evaluation of LLM agents, exploiting its history-independent property to construct a test that helps disambiguate between reasoning and extrapolation accounts.
    \item We build a reusable and parametrically adjustable experimental framework that can decouple interdependence strength from statistical structure, providing a template that can be adapted to analyze limitations in other recursive belief tasks. 
\end{itemize}

\section{Problem Formulation}\label{section:research_framework}

\subsection{Game-theoretic Model}

Consider an $n$-player simultaneous-move global game. Each agent $i\in\{1,\dots,n\}$ has a fixed private value $\theta_i$, drawn i.i.d. from a common-knowledge distribution $F(\cdot)$. In each round $t$, after observing a public cost $p_t$, all agents simultaneously and independently choose whether to participate ($a_i=1$) or not ($a_i=0$) in a public project with positive externalities. The total number of participants is denoted by $N_t=\sum_i a_i$. Agent $i$'s payoff is:

\begin{equation}
u_i(a_i, a_{-i}) = 
\begin{cases}
\theta_i + \beta N_t - p_t, & a_i=1,\\
0, & a_i=0,
\end{cases}
\label{equation:payoff}
\end{equation}

where $\beta>0$ captures the marginal external benefit. When $\beta=0$, the game reduces to an independent decision-making problem; as $\beta$ grows, payoffs become increasingly sensitive to beliefs about others' actions, which amplifies the role of higher‑order expectations and strategic interdependence in determining payoffs. Thus, as $\beta$ increases, payoffs become more sensitive to beliefs about others’ actions, amplifying the importance of higher-order expectations in equilibrium play.

\subsection{Rational Expectations Equilibrium}

We adopt the rational expectations equilibrium (REE) as a benchmark of rational play under full rationality. In a symmetric REE, all agents follow the same threshold strategy: there exists a cutoff $\theta^*(p_t)$ such that agent $i$ participates if and only if $\theta_i \ge \theta^*(p_t)$. The equilibrium requires consistency between beliefs and actual play: each agent's expectation of others' behavior exactly matches their actual behavior under that threshold strategy. We adopt proximity to the REE as a behavioral proxy. Our framework rests on the explicit premise that, under a broader bounded rationality perspective, a closer empirical alignment to the REE reflects deeper iterative reasoning.

Under the threshold strategy, for agent $i$, the participation probability of any other agent $j$ ($j \neq i$) is $\Pr(\theta_j \ge \theta^*) = 1 - F(\theta^*)$. Since the decisions of other agents are perceived as i.i.d. by $i$, $i$'s expectation of the number of participants excluding himself is $(n-1)(1 - F(\theta^*))$. Indifference for the marginal agent ( $\theta_i = \theta^*$) gives:

\begin{equation}
\theta^* + \beta \cdot \mathbb{E}_{-i}[N_t \mid \theta^*] = p_t,
\label{equation:indifference}
\end{equation}

where $\mathbb{E}_{-i}[N_t \mid \theta^*]$ is the marginal agent's expectation of the total number of participants. Since the agent himself participates ($a_i=1$) and each of the remaining $n-1$ agents participates with probability $1-F(\theta^*)$, we have

\begin{equation}
\mathbb{E}_{-i}[N_t \mid \theta^*] = 1 + (n-1)\bigl(1 - F(\theta^*)\bigr).
\label{equation:expectation_N}
\end{equation}

Substituting \eqref{equation:expectation_N} into \eqref{equation:indifference} yields the fixed-point equation determining $\theta^*(p_t)$:

\begin{equation}
\theta^* + \beta\Bigl[1 + (n-1)\bigl(1 - F(\theta^*)\bigr)\Bigr] = p_t.
\label{equation:fixedpoint}
\end{equation}

When $F$ satisfies suitable regularity conditions, \eqref{equation:fixedpoint} admits a unique solution $\theta^*(p_t)$. Appendix~\ref{appendix:equilibrium} provides the complete proof of existence and uniqueness.

Equation~\eqref{equation:fixedpoint} shows that the REE threshold is history‑independent. The strategy threshold $\theta^*(p_t)$ depends only on the current public signal $p_t$, the interdependence intensity $\beta$, and the private-value distribution $F$; it is independent of any statistical features of the historical price sequence $\{p_1,\dots,p_{t-1}\}$. Regardless of whether prices exhibit stationary fluctuations, a monotonic trend, or irregular jumps, as long as $p_t$ is the same, a rational agent's optimal strategy threshold remains unchanged. This follows from the rational‑expectations logic. Under the common-knowledge prior, all inferences about others' strategies are already encapsulated in $p_t$ and $F$, and the historical path provides no additional causal information. This history independence yields a testable restriction for distinguishing reasoning from extrapolation. 

\subsection{Research Hypotheses}

The framework translates the ICL mechanism question into two candidate mechanisms:

\begin{itemize}
    \item \textbf{H0 (Recursive Reasoning)}: Agents' decisions are primarily driven by strategic reasoning based on the current game structure; their behavior follows the reaction function characterized by the REE and is independent of historical statistical patterns.
    \item \textbf{H1 (Statistical Extrapolation)}: Agents' decisions are primarily driven by pattern matching and autoregressive extrapolation of the $(p, N)$ sequence within the context window; their behavior is strongly governed by the statistical structure of the sequence.
\end{itemize}

Manipulating the statistical structure of the price sequence in the interdependence game yields predictions that separate the two mechanisms:

\begin{itemize}
    \item If H0 dominates, reordering the same set of prices does not systematically affect decision quality, since the REE threshold depends only on $p_t$.
    \item If H1 dominates, decision quality is systematically higher under clear trends than under irregular or no trends, and this gap widens with $\beta$, due to amplification through strategic complementarity.
\end{itemize}

\section{Experimental Design}\label{section:experiment}

\subsection{Game Environment and Control Logic}

We instantiate the game from Section~\ref{section:research_framework} as a public goods game with $n=50$, private values $\theta_i\sim\mathcal{U}[0,49]$, and interdependence strength $\beta\in\{0.25,0.75\}$.

Each experimental session consists of $T=6$ decision rounds, with each round following a fixed protocol: the environment announces the current public cost $p_t$ and the interaction history up to the previous round; each agent independently decides whether to participate based on its own $\theta_i$ and the context; the environment then aggregates the total number of participants and returns individual payoffs.

The entire workflow is driven by the finite state machine shown in Figure~\ref{figure:finite_state_machine}, with a formal specification provided in Appendix~\ref{appendix:finite_state_machine}.

\begin{figure}[!htb]
    \centering
    \includegraphics[width=1.0\columnwidth]{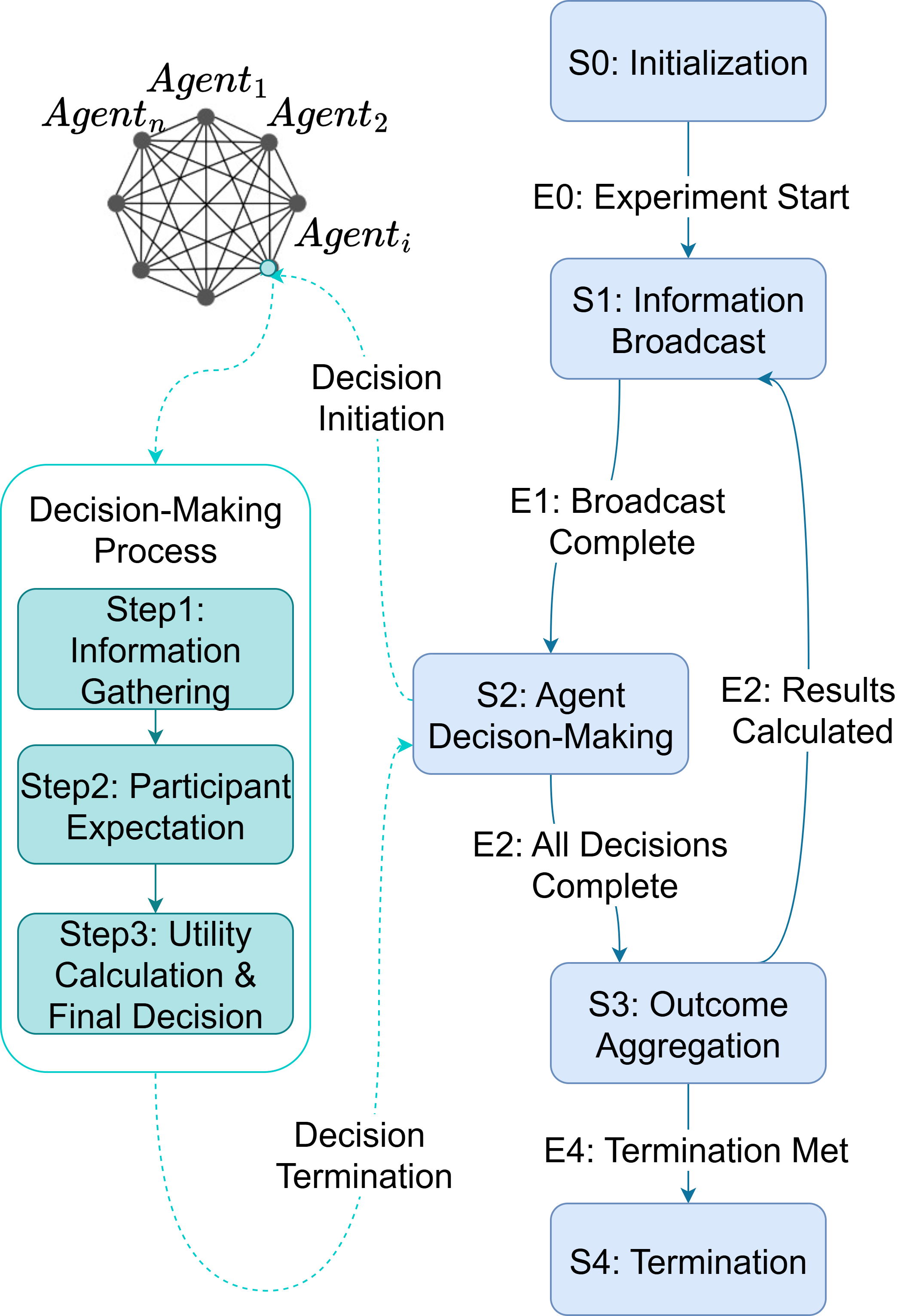}
    \caption{\textbf{Dynamic Feedback Loop for Agent Decision-Making.} The experimental framework operates as such a loop, where environmental signals drive autonomous agent decisions. Each iteration synchronizes public information broadcasting with internal belief updates and subsequent outcome aggregation, ensuring a rigorous causal trace from context to collective action.}
    \label{figure:finite_state_machine}
\end{figure}

\subsection{Price Sequence Structure}

To separate reasoning from extrapolation, we keep the fundamentals of each round fixed and only manipulate the temporal ordering of prices. We uniformly select $M=6$ price points to form the set $\mathcal{P}$ such that the corresponding equilibrium participation probabilities are sufficiently spread out.

Fixing $\mathcal{P}$, we change only the arrangement of prices across the $T$ rounds to construct two types of sequences:

\begin{itemize}
    \item \textbf{Monotonic sequences}: Prices change in a single direction, exhibiting a clear local trend;
    \item \textbf{Jump sequences}: Adjacent prices frequently switch between high and low values, forming no stable trend.
\end{itemize}

The inferential logic behind this manipulation is as follows: if agents rely on recursive reasoning, decisions under the same $p_t$ should not vary with the sequence type; if they depend on statistical extrapolation, decision quality in monotonic sequences will be significantly better than in jump sequences, and this difference will be further amplified by strategic complementarity when $\beta$ is large. Specific trajectories are provided in Appendix~\ref{appendix:price_sequences}.

\subsection{Context Structure}\label{subsection:context_window}

The historical information $\mathcal{H}_{i,t}^{(k)}$ available to agent $i$ in round $t$ is defined as the public signals, total participation numbers, and individual payoffs from the most recent $k$ rounds:

\begin{equation}
    \mathcal{H}_{i,t}^{(k)} = \big\{ (p_{t-\tau}, N_{t-\tau}, u_{i,t-\tau}) \big\}_{\tau=1}^{\min(k, t-1)},
    \label{equation:history_information}
\end{equation}

where $k \in \{0,3,6\}$. In the input implementation, $K$ rounds of history correspond to $c=2K+1$ messages, and the number of messages actually manipulated in the experiment is $c\in\{1,7,13\}$.

\subsection{Evaluation Metrics and Models}

To enable a unified metric of decision quality across different $\beta$, we follow the method in
Appendix~\ref{appendix:normalization} to determine lower and upper price bounds for each
$\beta$, and map the original public cost $p_t$ to $[0,5]$ via an affine transformation,
denoting the mapped price as $\tilde{p}_t$. Under this normalized scale, the rational
expectations equilibrium derived in Section~\ref{section:research_framework} collapses to
a $\beta$-independent linear benchmark:

\begin{equation}
    N_{\mathrm{eq}}(\tilde{p}) = -10\,\tilde{p} + 50.
    \label{equation:ree_benchmark}
\end{equation}

The derivation and the proof that this mapping preserves the history independence of REE
are provided in Appendix~\ref{appendix:equilibrium}.

For each experimental trajectory, we define the following three metrics to characterize the relationship between the collective behavior of agents and the equilibrium.

\paragraph{Equilibrium Deviation}

\begin{equation}
    \mathrm{ED} = \sqrt{\frac{1}{T}\sum_{t=1}^{T} \big( N_t - N_{\mathrm{eq}}(\tilde{p}_t) \big)^2 }.
    \label{equation:ED}
\end{equation}

$ED$ measures the deviation of actual participation from the equilibrium prediction.

\paragraph{Equilibrium Coefficient of Determination}

\begin{equation}
    R^2_{\mathrm{eq}} = 1 - \frac{\sum_{t=1}^{T} \big( N_t - N_{\mathrm{eq}}(\tilde{p}_t) \big)^2}{\sum_{t=1}^{T} \big( N_t - \bar{N} \big)^2}.
    \label{equation:R2_eq}
\end{equation}

where $\bar{N} = \frac{1}{T}\sum_{t=1}^{T} N_t$. $R^2_{\mathrm{eq}}$ captures how well the equilibrium line explains the shape of participation fluctuations.

\paragraph{Strategic Regression}

To further reveal the response rule that actually emerges from the collective interaction of agents, we perform ordinary least squares regression on the per‑round observations $(\tilde{p}_t, N_t)$:

\begin{equation}
    N_t = \alpha + \gamma\, \tilde{p}_t + \varepsilon_t.
    \label{equation:ols}
\end{equation}

Experiments are run on GPT-5 and Qwen3-Plus. Model configurations are detailed in Appendix~\ref{appendix:experimental_setup}. The main text reports representative results from GPT-5. Cross-model comparisons are provided in Appendix~\ref{appendix:results}.

\section{Experimental Results and Analysis}\label{section:results}

We report group decision-making behavior under three experimental conditions, which form a progressive diagnostic chain: first, we establish the baseline inference ability without any history; second, we examine the apparent benefit of ICL under a clear trend; finally, we test whether this benefit disappears when the trend is broken.

\subsection{Static Baseline}\label{subsection:static_results}

\begin{figure}[!htb]
    \centering
    \includegraphics[width=0.45\textwidth]{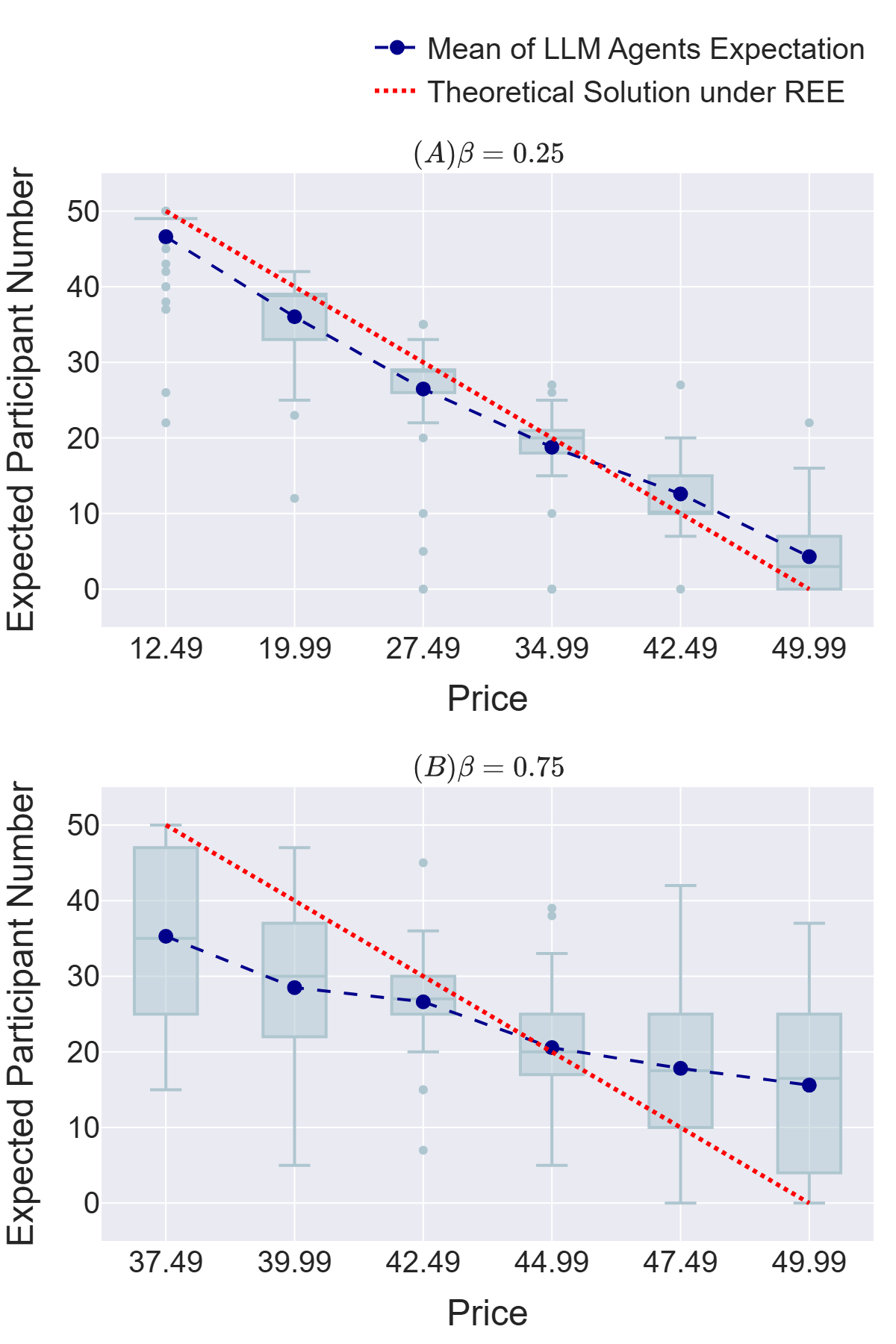}
    \caption{\textbf{Static baseline without interaction history.} Blue boxplots show the distribution of expected participation across six independent price points, with blue dashed line connecting group means and red dashed line representing the REE. The model exhibits systematic deviation from REE even without strategic interdependence: the response curve is flatter than equilibrium, overestimating participation at high prices and underestimating at low prices. The deviation intensifies sharply when the interdependence parameter $\beta$ increases from $0.25$ to $0.75$, revealing a fundamental limitation in recursive belief reasoning.}
    \label{figure:gpt_5_static_results}
\end{figure}

When no interaction history is provided, agents make participation decisions based solely on the current price $p_t$ and their private value $\theta_i$.
Figure~\ref{figure:gpt_5_static_results} summarizes the group expectation distributions across six independent price points. This baseline characterizes the model's basic decision-making pattern without any context and serves as a reference for measuring the net contribution of ICL.

Expected participation decreases with price, consistent with theory, but the curve is flatter than REE, leading to systematic overestimation at high prices and underestimation at low prices.
Table~\ref{table:metrics_gpt_5} shows that increasing $\beta$ nearly doubles $ED$ and cuts $R^2_{\mathrm{eq}}$ by more than half, while box widths grow, confirming that stronger interdependence amplifies deviations and reduces consistency.
Thus, even without history, the model shows systematic limitations in recursive belief reasoning.

We visualize only increasing and converging trajectories here. Full results and metrics are in Appendix~\ref{appendix:supplementary_results}.

\subsection{Monotonic Sequences}\label{subsection:dynamic_monotonic}

\begin{figure*}[!htb]
    \centering
    \includegraphics[width=1.0\linewidth]{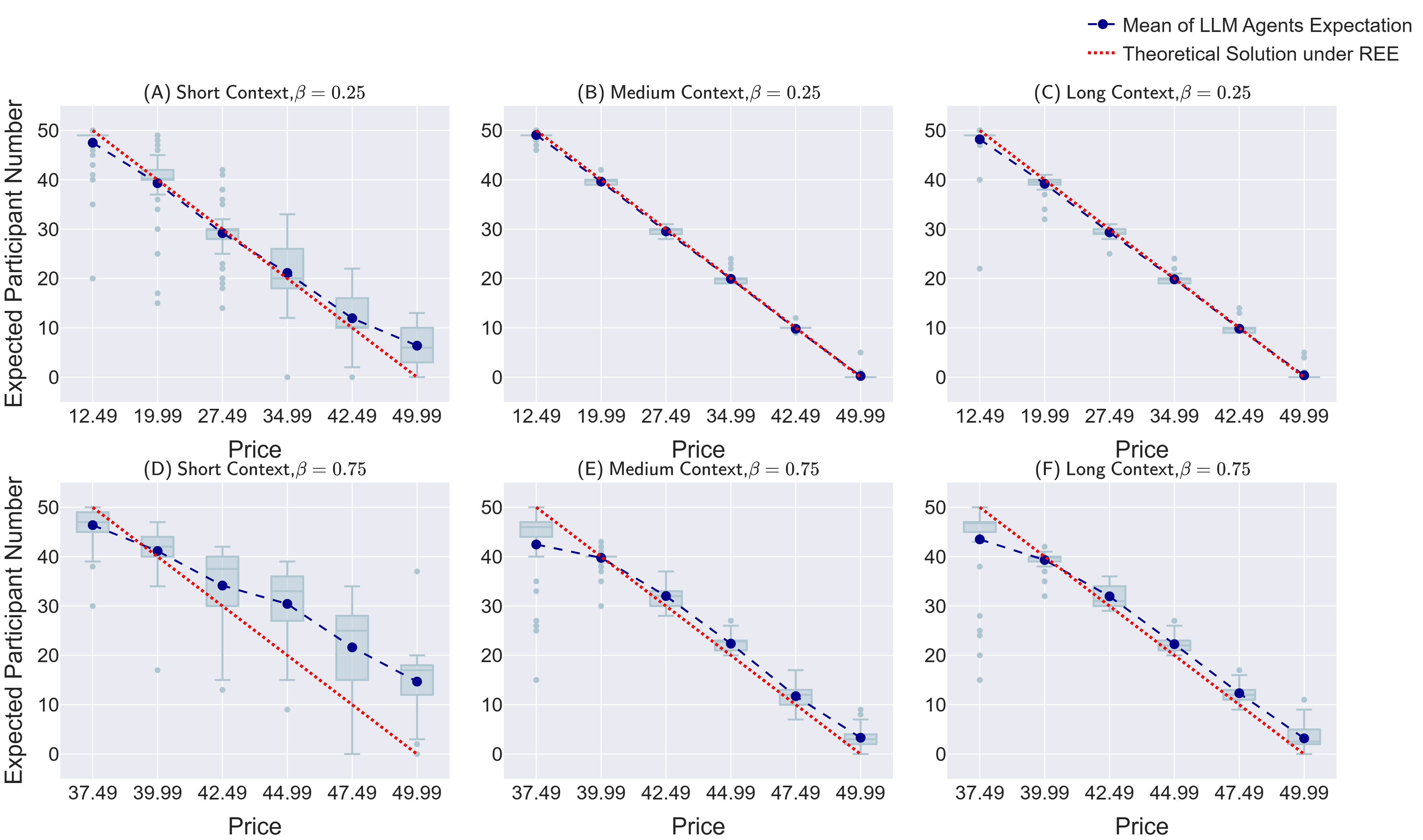}
    \caption{\textbf{Monotonic price trajectory.} The columns show context window lengths $k=0,3,6$; top row $\beta=0.25$, bottom row $\beta=0.75$. When a clear temporal trend is present, providing interaction history dramatically shrinks prediction spread and draws the group mean toward the REE line. However, the improvement exhibits diminishing returns, fails to eliminate the systematic bias, and remains highly sensitive to $\beta$. The pattern is consistent with statistical trend extrapolation rather than recursive reasoning.}
    \label{figure:gpt_5_increasing_price_results}
\end{figure*}

We introduce historical information, beginning with monotonic price sequences. As a representative case, Figure~\ref{figure:gpt_5_increasing_price_results} shows an increasing trajectory under expanding context lengths $k=0,3,6$.

\begin{figure*}[!htb]
    \centering
    \includegraphics[width=1.0\linewidth]{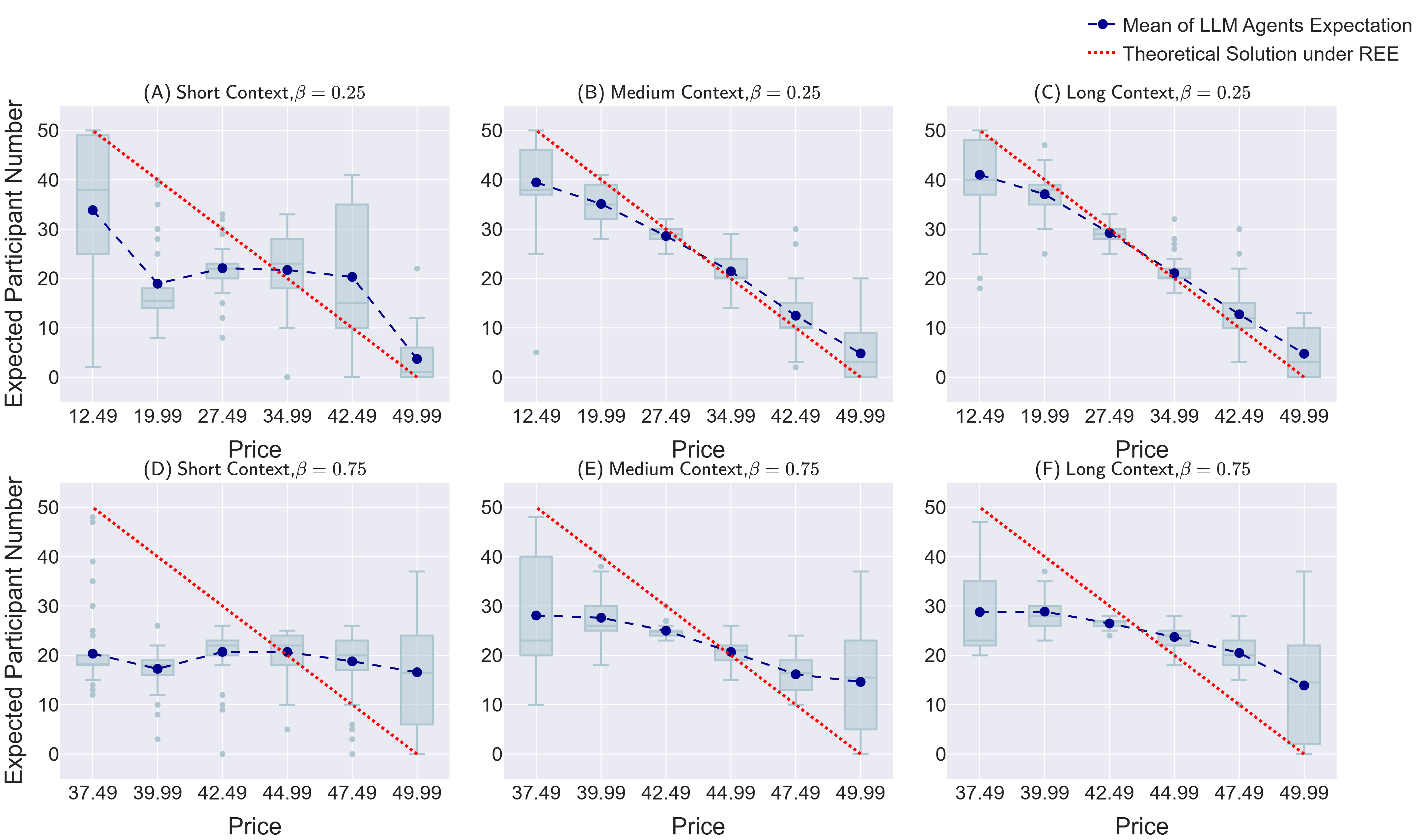}
    \caption{\textbf{Jump price trajectory.} The price sequence oscillates without a sustained trend. Unlike the monotonic case, expanding the context window provides virtually no systematic improvement: prediction spreads remain wide, group means stay far from REE, and at high $\beta$ the model performs no better than or even worse than the no-history baseline. The disappearance of ICL’s benefit precisely when the statistical trend is removed aligns closely with the extrapolation hypothesis.}
    \label{figure:gpt_5_converging_price_results}
\end{figure*}

With only three rounds of history, expectation spread shrinks and group means move closer to REE (Table~\ref{table:metrics_gpt_5}: for $\beta=0.25$, $ED$ drops from 6.18 to 0.84; for $\beta=0.75$, from 11.14 to 5.60). However, the improvement (1) shows diminishing returns ($ED$ changes little from $k=3$ to $k=6$), (2) does not eliminate systematic bias, and (3) remains sensitive to $\beta$ ($ED$ is much higher for $\beta=0.75$). Regression slopes (Table~\ref{table:partial_regression_summary_gpt_5}) approach $-10$ only for $\beta=0.25$, staying too flat for $\beta=0.75$. Hence, the monotonic-trend benefit is better explained as statistical extrapolation than as recursive reasoning.

\subsection{Jump Sequences}\label{subsection:dynamic_non_monotonic}

If the above interpretation is correct, the behavioral improvement from ICL should disappear when the statistical regularity of the historical sequence is disrupted.
The non-monotonic jump sequence is designed precisely as a diagnostic condition to test this inference: prices switch frequently between high and low values, offering no stable direction for autoregressive extrapolation.
Figure~\ref{figure:gpt_5_converging_price_results} presents the results under this condition.

In this setting, longer context provides no systematic benefit. For $\beta=0.25$, $ED$ remains above 6.5 and $R^2_{\mathrm{eq}}$ peaks at 0.854; for $\beta=0.75$, $ED$ exceeds 13.88, $R^2_{\mathrm{eq}}$ becomes negative at $k=0$, and slopes stay far below $-10$ (e.g., $\gamma=-3.03$ at $k=6$). Spread does not narrow, and performance sometimes worsens relative to the no-history baseline. The failure is amplified at higher $\beta$. This matches extrapolation: without a learnable trend, ICL yields no improvement, especially under high interdependence.

\subsection{Quantitative Analysis Summary}\label{subsection:quantitative_analysis}

To present a complete picture of group performance under all experimental conditions, Table~\ref{table:metrics_gpt_5} summarizes the equilibrium deviation and equilibrium coefficient of determination for five price trajectories, and Table~\ref{table:regression_summary_gpt_5} reports the corresponding regression intercept $\alpha$, slope $\gamma$, and model fit.

\begin{table}[!htb]
\centering
\small
\caption{Equilibrium deviation and equilibrium determination coefficient across all trajectories. }
\label{table:metrics_gpt_5}
\begin{tabular}{ccccc}
\toprule
$\bm{\mathbf{\beta}}$ & \textbf{Trajectory} & $\bm{\mathbf{k}}$ & $\bm{\mathrm{ED}}$ & $\bm{\mathrm{R}^2_{eq}}$ \\
\midrule
\multirow{13}{*}{0.25} & Static & -- & 6.712 & 0.846 \\
\cmidrule{2-5}
 & \multirow{3}{*}{Decreasing} & 0 & 9.031 & 0.720 \\
 &  & 3 & 2.124 & 0.985 \\
 &  & 6 & 2.486 & 0.979 \\
\cmidrule{2-5}
 & \multirow{3}{*}{Increasing} & 0 & 6.180 & 0.869 \\
 &  & 3 & 0.843 & 0.998 \\
 &  & 6 & 2.154 & 0.906 \\
\cmidrule{2-5}
 & \multirow{3}{*}{Converging} & 0 & 15.912 & 0.132 \\
 &  & 3 & 7.313 & 0.817 \\
 &  & 6 & 6.534 & 0.854 \\
\cmidrule{2-5}
 & \multirow{3}{*}{Diverging} & 0 & 17.211 & -0.016 \\
 &  & 3 & 3.292 & 0.963 \\
 &  & 6 & 4.301 & 0.937 \\
\midrule
\multirow{13}{*}{0.75} & Static & -- & 13.979 & 0.330 \\
\cmidrule{2-5}
 & \multirow{3}{*}{Decreasing} & 0 & 17.203 & -0.015 \\
 &  & 3 & 6.405 & 0.859 \\
 &  & 6 & 5.026 & 0.913 \\
\cmidrule{2-5}
 & \multirow{3}{*}{Increasing} & 0 & 11.136 & 0.575 \\
 &  & 3 & 5.596 & 0.893 \\
 &  & 6 & 5.242 & 0.906 \\
\cmidrule{2-5}
 & \multirow{3}{*}{Converging} & 0 & 18.731 & -0.203 \\
 &  & 3 & 14.179 & 0.311 \\
 &  & 6 & 13.884 & 0.339 \\
\cmidrule{2-5}
 & \multirow{3}{*}{Diverging} & 0 & 18.713 & -0.201 \\
 &  & 3 & 8.779 & 0.736 \\
 &  & 6 & 9.120 & 0.715 \\
\bottomrule
\end{tabular}
\end{table}

\begin{table}[!htb]
\centering
\small
\caption{OLS regression of observed participation on mapped price.}
\label{table:partial_regression_summary_gpt_5}
\begin{tabular}{cccccc}
\toprule
$\bm{\beta}$ & \textbf{Trajectory} & $\bm{k}$ & $\bm{\alpha}$ & $\bm{\gamma}$ & $\bm{R^2 / \bar{R}^2}$ \\
\midrule
\multirow{13}{*}{0.25} & Static & -- & 44.81 & -8.27 & 0.849 / 0.848 \\
\cmidrule{2-6}
 & \multirow{3}{*}{Decreasing} & 0 & 39.67 & -7.78 & 0.799 / 0.799 \\
 &                             & 3 & 49.16 & -9.78 & 0.985 / 0.985 \\
 &                             & 6 & 48.85 & -9.67 & 0.979 / 0.979 \\
\cmidrule{2-6}
 & \multirow{3}{*}{Increasing} & 0 & 47.03 & -8.45 & 0.873 / 0.872 \\
 &                             & 3 & 49.21 & -9.81 & 0.998 / 0.998 \\
 &                             & 6 & 48.51 & -9.62 & 0.986 / 0.986 \\
\cmidrule{2-6}
 & \multirow{3}{*}{Converging} & 0 & 30.60 & -4.20 & 0.282 / 0.280 \\
 &                             & 3 & 41.38 & -7.10 & 0.844 / 0.844 \\
 &                             & 6 & 43.04 & -7.50 & 0.873 / 0.872 \\
\cmidrule{2-6}
 & \multirow{3}{*}{Diverging}  & 0 & 26.25 & -2.29 & 0.130 / 0.127 \\
 &                             & 3 & 47.49 & -9.20 & 0.966 / 0.966 \\
 &                             & 6 & 47.33 & -9.19 & 0.938 / 0.938 \\
\midrule
\multirow{13}{*}{0.75} & Static & -- & 33.81 & -3.90 & 0.340 / 0.338 \\
\cmidrule{2-6}
 & \multirow{3}{*}{Decreasing} & 0 & 23.21 & -4.04 & 0.483 / 0.481 \\
 &                             & 3 & 43.45 & -8.26 & 0.879 / 0.879 \\
 &                             & 6 & 44.44 & -8.52 & 0.932 / 0.932 \\
\cmidrule{2-6}
 & \multirow{3}{*}{Increasing} & 0 & 47.17 & -6.31 & 0.728 / 0.727 \\
 &                             & 3 & 45.95 & -8.27 & 0.899 / 0.898 \\
 &                             & 6 & 46.30 & -8.35 & 0.913 / 0.913 \\
\cmidrule{2-6}
 & \multirow{3}{*}{Converging} & 0 & 20.08 & -0.41 & 0.010 / 0.007 \\
 &                             & 3 & 29.60 & -3.03 & 0.347 / 0.344 \\
 &                             & 6 & 31.01 & -2.92 & 0.356 / 0.354 \\
\cmidrule{2-6}
 & \multirow{3}{*}{Diverging}  & 0 & 22.65 &  0.28 & 0.006 / 0.003 \\
 &                             & 3 & 40.75 & -6.09 & 0.771 / 0.770 \\
 &                             & 6 & 40.50 & -5.85 & 0.757 / 0.756 \\
\bottomrule
\end{tabular}
\end{table}

The data in these tables are highly consistent with the analysis of the increasing and jump sequences in the previous three subsections: when the price sequence has an extrapolable monotonic trend, introducing history substantially reduces $ED$ and brings the regression slope close to the theoretical value of $-10$; in sequences without a trend, $ED$ remains high, $R^2_{\mathrm{eq}}$ improves only marginally or becomes negative, and $\gamma$ deviates severely from equilibrium. The interdependence strength $\beta$ consistently acts as an amplifier: for $\beta=0.75$, decision quality across all trajectories is worse than under the corresponding $\beta=0.25$ conditions. More detailed statistical test results are provided in the Appendix~\ref{appendix:full_sample_analysis}.

\section{Related Work}

\subsection{Mechanisms of In-Context Learning in Decision-Making}

In-context learning enables LLMs to adapt to downstream tasks without parameter updates~\citep{brown2020, wei2022}.
It has been extended to sequential decision-making, where models improve actions by incorporating interaction histories and feedback into the context~\citep{song2025reward, xia2025, chen2025retrospective}.
\citet{song2025reward} show that LLMs can maximize scalar rewards over multiple prompting rounds, a phenomenon termed in-context reinforcement learning.
\citet{chen2025retrospective} demonstrate that LLMs transform sparse feedback into dense training signals via retrospective ICL.

A fundamental question remains: how does ICL produce behavioral improvement? One view interprets ICL as implicit reasoning: \citet{xie2022} propose that models perform Bayesian inference over latent concepts, a perspective supported by recent theoretical analyses~\citep{wakayama2025bayesian} and empirical studies showing that LLMs update their predictions in a Bayes-consistent manner given sufficient demonstrations~\citep{gupta2025coin}. Others show transformers can internally implement gradient descent on in-context examples~\citep{von2023transformers, akyurek2023learning} and even simulate multi-step optimization of deep neural networks~\citep{wu2025deeplearning}. An alternative view stresses statistical pattern matching: \citet{olsson2022context} identify induction heads that copy tokens, arguing this mechanism underlies general ICL ability; subsequent ablation studies confirm that disabling induction heads substantially degrades few-shot ICL performance~\citep{crosbie2025induction}.

\subsection{Recursive Belief Reasoning and Multi-Agent Games}

Strategic interaction requires recursive belief reasoning because optimal actions depend on expectations about others' actions.
Behavioral game theory models such as level-$k$~\citep{nagel1995unraveling, stahl1995players} and cognitive hierarchy~\citep{camerer2004cognitive} characterize individuals' limited depths of strategic thinking.
The beauty contest game~\citep{nagel1995unraveling} has become a canonical paradigm for measuring recursive reasoning in human populations.

Recent work evaluates LLMs' strategic reasoning through game-theoretic lenses.
\citet{kempinski2025game} guide LLMs to iteratively refine actions in self-play, resembling cognitive hierarchy.
\citet{Trencsenyi2025ApproximatingHS} employ hypergames to assess recursive reasoning in one-shot beauty contest games.
\citet{yuan2026marshal} develop MARSHAL, an RL framework incentivizing multi-agent reasoning via self-play.
These efforts focus on whether LLMs exhibit strategic reasoning, without isolating the underlying mechanism.
Systematically manipulating interdependence strength and feedback statistics offers a way to probe the mechanistic boundaries of ICL.

\subsection{Global Games and Rational Expectations Equilibrium}

Global games~\citep{carlsson1993global, morris2003global} analyze coordination under incomplete information, where agents receive noisy private signals and equilibrium takes the form of threshold strategies.
A key property is uniqueness even when complete-information games admit multiple equilibria, facilitating empirical evaluation.
Rational expectations equilibrium \citep{John1961, lucas1972expectations} imposes consistency between beliefs and actual distributions of actions, yielding a benchmark of rational play.
A crucial feature of REE is its history independence: the equilibrium strategy depends only on contemporaneous fundamentals, not on historical realizations.
This property makes REE an ideal reference for distinguishing reasoning-based from extrapolation-based behavior.

Global games and REE have been extensively studied in economics~\citep{Harsanyi2004, angeletos2007dynamic} and recently applied to analyze LLM behavior in social dilemmas~\citep{liang2025everyone}.
However, exploiting REE's history independence as a diagnostic tool for probing the mechanism of ICL in multi-agent settings remains underexplored.

\subsection{LLM Agents in Game-Theoretic Experiments}

Game-theoretic paradigms are increasingly used to evaluate LLM agents' strategic behavior.
\citet{liang2025everyone} design sequential public goods games to incentivize cooperation in multi-LLM systems.
\citet{piedrahita2025corruptedreasoningreasoninglanguage} adapt a public goods game with institutional choice, finding that reasoning-focused LLMs paradoxically become free-riders.
\citet{huynh2025understandingllmagentbehaviours} apply the FAIRGAME framework to repeated social dilemmas, revealing systematic cooperation biases across models and languages.

These studies characterize cooperative or competitive tendencies but do not disentangle the mechanisms driving ICL-based behavioral change.
Disentangling reasoning from extrapolation by jointly manipulating history structure and interdependence strength provides a complementary diagnostic approach, building on the history‑independence property of REE.

\section{Conclusion}

In this paper, we investigated whether in-context learning in LLM agents operating in multi-agent interdependent settings is better characterized as recursive belief reasoning or as statistical extrapolation of observed patterns. Using a repeated public goods game with a history-independent rational expectations equilibrium benchmark, we manipulated the statistical structure of the feedback sequence to separate these two candidate mechanisms.

Our results show that the improvements associated with in-context learning are closely tied to the presence of a clear temporal trend in the context. When the trend is removed, the advantage of additional context largely disappears, and this pattern becomes more pronounced at higher levels of interdependence. These observations are consistent with the interpretation that in-context learning in our setting relies primarily on statistical extrapolation rather than on recursive belief updating toward equilibrium play.

We hope that the framework introduced here, based on the history-independence of rational expectations equilibrium, will offer a useful diagnostic tool for studying how and when LLM agents engage in recursive reasoning. Future work could extend this framework to probe finer-grained strategic behavior and to test whether interventions can shift behavior toward equilibrium-consistent play.

\section{Limitations}

Our study has several limitations that suggest natural directions for future work. First, we evaluated only two model families, so the observed patterns may not generalize to all LLM architectures or scales. Second, the public goods game, while effective for controlled manipulation, captures one specific form of strategic interdependence, and different game structures could in principle elicit different behaviors. Third, we explored relatively short context windows, leaving open the possibility that substantially longer histories might alter the balance between extrapolation and reasoning. Fourth, the analysis is purely behavioral; connecting these patterns to internal model mechanisms remains an open challenge. Finally, we employed a fixed prompt format, and sensitivity to linguistic framing was not assessed. None of these issues undermine the main finding, but addressing them will build a more complete picture of how in-context learning operates in strategic multi-agent settings.

\bibliography{contents/references/main_references}

\appendix

\section{Equilibrium Derivation and Construction of Normalized Benchmark}\label{appendix:equilibrium}

This appendix provides the detailed derivation of the REE described in Section~\ref{section:research_framework} of the main paper, and explains the construction logic of the normalized evaluation benchmark.

\subsection{General Existence and Uniqueness Condition}\label{appendix:solution_existence}

Consider the fixed-point equation from equation~\eqref{equation:fixedpoint} in the main paper:

\begin{equation}
    \theta^* + \beta\Bigl[1 + (n-1)\bigl(1 - F(\theta^*)\bigr)\Bigr] = p_t,
    \label{appendix:equation_fixpoint}
\end{equation}

where $F(\cdot)$ is the cumulative distribution function of the private value $\theta_i$. Define the function

\begin{equation}
    G(\theta) = \theta + \beta\bigl[1 + (n-1)(1 - F(\theta))\bigr].
\end{equation}

Since $F$ is continuously differentiable and $f(\theta)\ge 0$, we have:

\begin{equation}
    G'(\theta) = 1 - \beta (n-1) f(\theta).
\end{equation}

As long as $\beta (n-1) f(\theta)<1$ holds for all $\theta$, $G$ is strictly increasing, and thus for any $p_t$ there exists a unique solution $\theta^*(p_t)$. Under the experimental settings of this paper ($\theta_i$ uniformly distributed, $\beta\in\{0.25,0.75\}$, $n=50$), this condition is naturally satisfied, and an equilibrium always exists and is unique.

\subsection{Explicit Equilibrium Solution under Uniform Distribution}\label{appendix:uniform_distribution_solution}

Assume that the private values $\theta_i$ are i.i.d. uniformly distributed on the interval $[a,b]$, i.e., $F(\theta)=(\theta-a)/(b-a)$. Substituting into \eqref{appendix:equation_fixpoint} and solving for $\theta^*$, we obtain the explicit expression for the equilibrium threshold:

\begin{equation}
    \theta^*(p) = \frac{p - \beta - \frac{\beta (n-1)b}{b - a}}{1 - \frac{\beta (n-1)}{b - a}}.
    \label{appendix:theta_star}
\end{equation}

The participation probability is $1 - F(\theta^*(p))$, so the equilibrium total number of participants is

\begin{equation}
    N_{\mathrm{eq}}(p) = n \cdot \frac{b + \beta - p}{(b - a) - \beta (n-1)}.
    \label{appendix:Neq_general}
\end{equation}

Thus, under the uniform distribution, the equilibrium number of participants is a linear function of the public price $p$, with slope and intercept depending on $\beta$ and the support of the distribution.

\subsection{Normalization Mapping and Unified Evaluation Scale}\label{appendix:normalization}

To measure decision quality under different $\beta$ values on a unified scale, we determine a price interval $[p_{\min}(\beta), p_{\max}(\beta)]$ for each $\beta$ and map it to $\tilde{p}\in[0,5]$ via the affine mapping:

\begin{equation}
    \tilde{p} = 5 \cdot \frac{p - p_{\min}(\beta)}{p_{\max}(\beta) - p_{\min}(\beta)}.
    \label{appendix:affine_map}
\end{equation}

The lower and upper bounds of the mapping are determined by inverting the equilibrium relation \eqref{appendix:Neq_general}:

\begin{itemize}
    \item $N_{\mathrm{eq}}=n$ corresponds to $p_{\min}(\beta)$;
    \item $N_{\mathrm{eq}}=0$ corresponds to $p_{\max}(\beta)$.
\end{itemize}

Thus, regardless of the value of $\beta$, the normalized equilibrium number of participants follows the same linear benchmark:

\begin{equation}
    N_{\mathrm{eq}}(\tilde{p}) = -10\tilde{p} + n.
    \label{appendix:unified_benchmark}
\end{equation}

In the experiments $n=50$, which is equation \eqref{equation:ree_benchmark} in the main paper.

This normalization is a deterministic transformation performed independently for each period and introduces no cross-period correlation, so the history-independence of the REE is fully preserved in the $(\tilde{p}, N)$ space. This guarantees the logical foundation for distinguishing recursive reasoning from statistical extrapolation by manipulating the ordering of price sequences in the main paper. It should be noted that the same $\tilde{p}$ value under different $\beta$ corresponds to different original prices $p$. Therefore, experimental inference relies on within-group effects of sequence type and its interaction with $\beta$, rather than on direct cross-$\beta$ comparisons of absolute deviations.

In summary, this appendix provides the theoretical guarantee of equilibrium existence and uniqueness, and on this basis constructs a standardized evaluation framework that preserves both theoretical identification power and experimental convenience.

\section{Detailed Results}\label{appendix:detailed_results}

\subsection{Supplementary Results}\label{appendix:supplementary_results}

\begin{figure*}[!htb]
    \centering
    \includegraphics[width=1.0\linewidth]{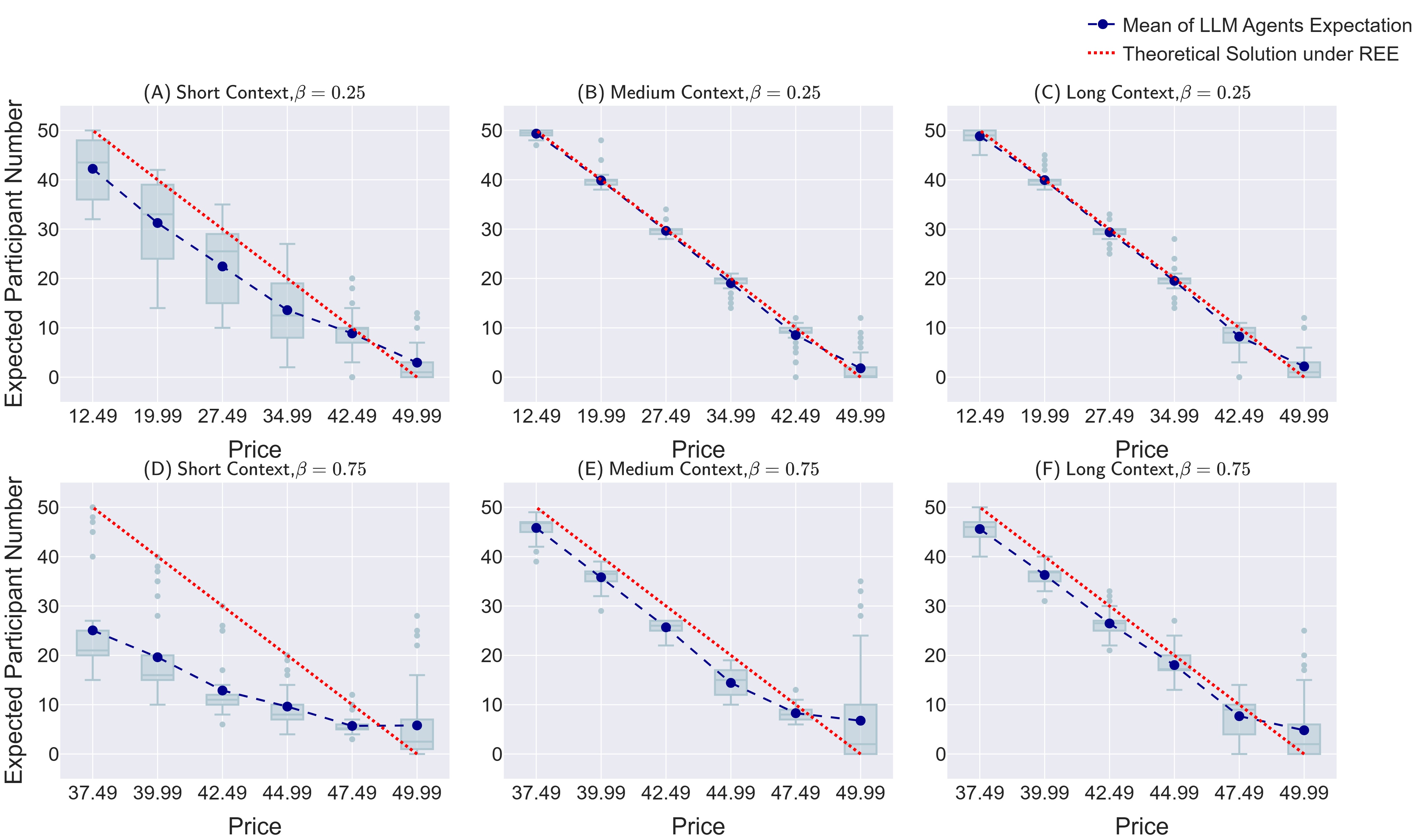}
    \caption{GPT-5: Monotonic decreasing price trajectory.}
    \label{figure:gpt_5_decreasing_price_results}
\end{figure*}

\begin{figure*}[!htb]
    \centering
    \includegraphics[width=1.0\linewidth]{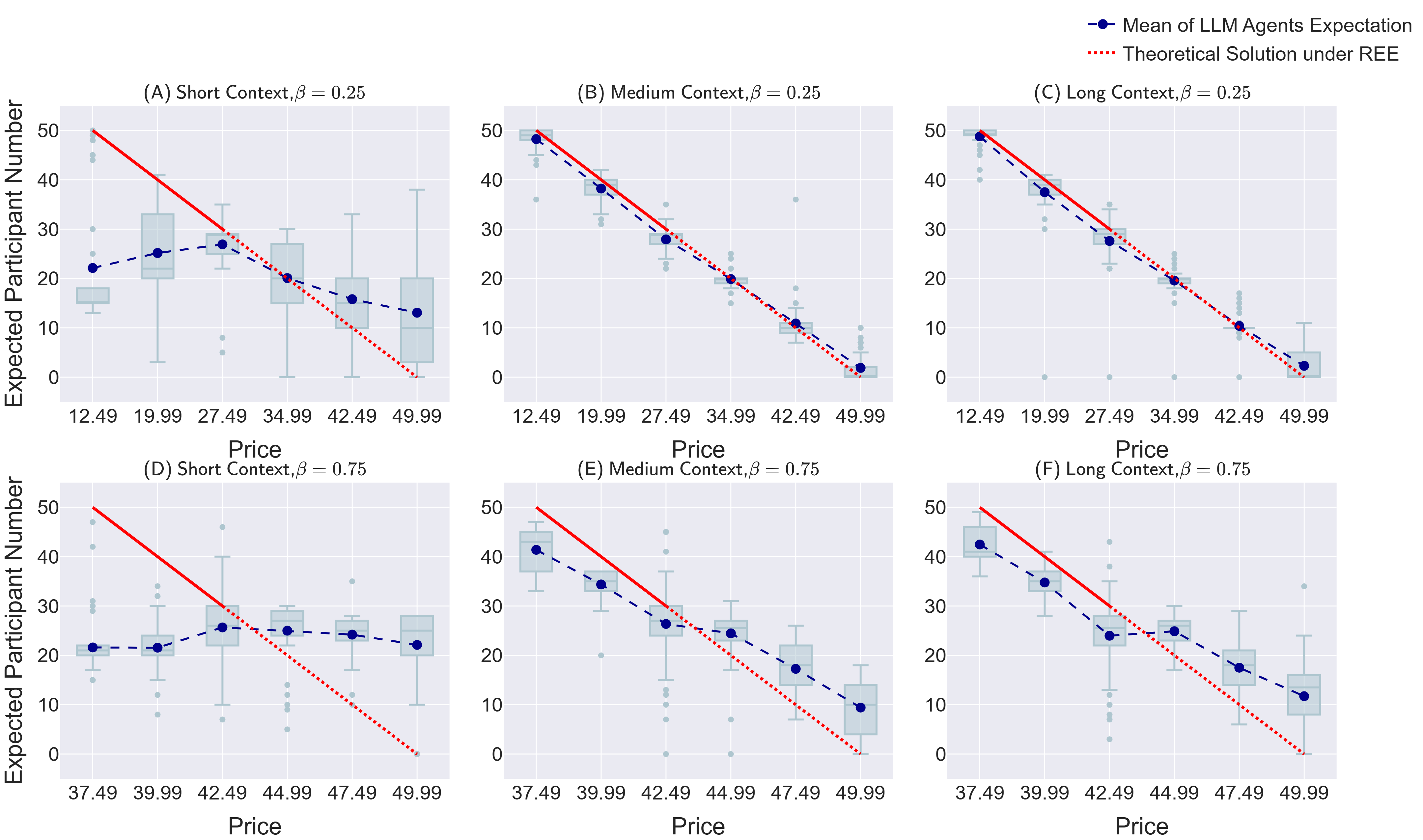}
    \caption{GPT-5: Jump diverging price trajectory.}
    \label{figure:gpt_5_diverging_price_results}
\end{figure*}

This section supplements the main text with all the details and complete experimental results for Section~\ref{section:results}. Figure~\ref{figure:gpt_5_decreasing_price_results} illustrates the decreasing trajectory under monotonic sequences, Figure~\ref{figure:gpt_5_diverging_price_results} illustrates the diverging trajectory under jump sequences, and Table~\ref{table:regression_summary_gpt_5} presents the full regression analysis results.

\begin{table*}[htb]
\centering
\small
\caption{GPT-5: OLS regression of observed participation on mapped price.}
\label{table:regression_summary_gpt_5}
\begin{tabular}{cc cccc}
\toprule
$\bm{\mathbf{\beta}}$ & \textbf{Trajectory} & $\bm{\mathbf{k}}$ & $\bm{\alpha}$ (t) & $\bm{\gamma}$ (t) & $\bm{\mathbf{R^2 / \bar{R}^2}}$ \\
\midrule
\multirow{13}{*}{0.25} & Static & -- & 44.81$^{***}$ (73.16) & -8.27$^{***}$ (-40.88) & 0.849 / 0.848 \\
\cmidrule{2-6}
 & \multirow{3}{*}{Decreasing} & 0 & 39.67$^{***}$ (58.03) & -7.78$^{***}$ (-34.47) & 0.799 / 0.799 \\
 &  & 3 & 49.16$^{***}$ (231.31) & -9.78$^{***}$ (-139.35) & 0.985 / 0.985 \\
 &  & 6 & 48.85$^{***}$ (198.30) & -9.67$^{***}$ (-118.82) & 0.979 / 0.979 \\
\cmidrule{2-6}
 & \multirow{3}{*}{Increasing} & 0 & 47.03$^{***}$ (83.09) & -8.45$^{***}$ (-45.20) & 0.873 / 0.872 \\
 &  & 3 & 49.21$^{***}$ (672.80) & -9.81$^{***}$ (-405.88) & 0.998 / 0.998 \\
 &  & 6 & 48.51$^{***}$ (238.39) & -9.62$^{***}$ (-143.13) & 0.986 / 0.986 \\
\cmidrule{2-6}
 & \multirow{3}{*}{Converging} & 0 & 30.60$^{***}$ (26.03) & -4.20$^{***}$ (-10.82) & 0.282 / 0.280 \\
 &  & 3 & 41.38$^{***}$ (77.47) & -7.10$^{***}$ (-40.22) & 0.844 / 0.844 \\
 &  & 6 & 43.04$^{***}$ (85.69) & -7.50$^{***}$ (-45.19) & 0.873 / 0.872 \\
\cmidrule{2-6}
 & \multirow{3}{*}{Diverging} & 0 & 26.25$^{***}$ (25.21) & -2.29$^{***}$ (-6.66) & 0.130 / 0.127 \\
 &  & 3 & 47.49$^{***}$ (156.75) & -9.20$^{***}$ (-91.89) & 0.966 / 0.966 \\
 &  & 6 & 47.33$^{***}$ (114.59) & -9.19$^{***}$ (-67.38) & 0.938 / 0.938 \\
\midrule
\multirow{13}{*}{0.75} & Static & -- & 33.81$^{***}$ (35.50) & -3.90$^{***}$ (-12.40) & 0.340 / 0.338 \\
\cmidrule{2-6}
 & \multirow{3}{*}{Decreasing} & 0 & 23.21$^{***}$ (31.65) & -4.04$^{***}$ (-16.67) & 0.483 / 0.481 \\
 &  & 3 & 43.45$^{***}$ (80.89) & -8.26$^{***}$ (-46.57) & 0.879 / 0.879 \\
 &  & 6 & 44.44$^{***}$ (110.25) & -8.52$^{***}$ (-63.98) & 0.932 / 0.932 \\
\cmidrule{2-6}
 & \multirow{3}{*}{Increasing} & 0 & 47.17$^{***}$ (69.77) & -6.31$^{***}$ (-28.25) & 0.728 / 0.727 \\
 &  & 3 & 45.95$^{***}$ (94.23) & -8.27$^{***}$ (-51.36) & 0.899 / 0.898 \\
 &  & 6 & 46.30$^{***}$ (102.44) & -8.35$^{***}$ (-55.94) & 0.913 / 0.913 \\
\cmidrule{2-6}
 & \multirow{3}{*}{Converging} & 0 & 20.08$^{***}$ (28.47) & -0.41 (-1.75) & 0.010 / 0.007 \\
 &  & 3 & 29.60$^{***}$ (40.59) & -3.03$^{***}$ (-12.57) & 0.347 / 0.344 \\
 &  & 6 & 31.01$^{***}$ (45.02) & -2.92$^{***}$ (-12.85) & 0.356 / 0.354 \\
\cmidrule{2-6}
 & \multirow{3}{*}{Diverging} & 0 & 22.65$^{***}$ (35.25) & 0.28 (1.33) & 0.006 / 0.003 \\
 &  & 3 & 40.75$^{***}$ (70.05) & -6.09$^{***}$ (-31.67) & 0.771 / 0.770 \\
 &  & 6 & 40.50$^{***}$ (69.66) & -5.85$^{***}$ (-30.44) & 0.757 / 0.756 \\
\bottomrule
\multicolumn{6}{l}{\footnotesize Note: $t$-statistics in parentheses. ${}^{***}p<0.001$, ${}^{**}p<0.01$, ${}^{*}p<0.05$.} \\
\end{tabular}
\end{table*}

\subsection{Full Sample Analysis}\label{appendix:full_sample_analysis}

The previous analysis evaluated agents’ collective deviations from REE using group-level $ED$ and $R^2_{\mathrm{eq}}$. We now move one step deeper and ask: how do individual agents deviate from the REE benchmark, and what systematic factors possibly drive these deviations? To capture this deviation, we define: 

\begin{equation}
Y \;=\; \hat{y}(p) - y_{\text{REE}}(p),
\end{equation}

where $\hat{y}(p)$ is the stated expectation of the number of participants at price $p$, and $y_{\text{REE}}(p)$ is the theoretical prediction under the same price. By construction, $Y=0$ should always hold for fully rational economic agents.

The estimation sample contains $7{,}800$ observations from the full-factorial experimental design. The regressors include the posted price ($p$), the interdependence intensity ($\beta \in \{0,1\}$, indicating weak or strong), the private value of the agent ($\theta$), and the accessible context window ($k \in \{0,0.5,1\}$, corresponding to 1, 7, and 13 rounds of history used previously, respectively). Because $Y$ can take both positive and negative values and exhibits heavy tails, we apply a Yeo–Johnson transformation to stabilize inference. All regressions include price-path fixed effects (static, increasing, decreasing, converging, and diverging trajectories).

\begin{equation}
\begin{split}
\text{Model 0:}\quad 
Y &= \beta_0 + \beta_1\times p + \beta_2 \times \beta \\
  & \quad + \beta_3 \times \theta \\
  & \quad+ \boldsymbol{\gamma} \mathbf{FE} + \varepsilon,
\end{split}
\end{equation}

\begin{equation}
\begin{split}
\text{Model 1:}\quad 
Y &= \beta_0 + \beta_1 \times p + \beta_2 \times \beta \\
  & \quad + \beta_3 \times \theta + \beta_4 \times k \\
  & \quad + \boldsymbol{\gamma} \mathbf{FE} + \varepsilon,
\end{split}
\end{equation}

\begin{equation}
\begin{split}
\text{Model 2:}\quad 
Y &= \beta_0 + \beta_1 \times p + \beta_2 \times \beta \\
  & \quad + \beta_3 \times \theta + \beta_4 \times k \\
  & \quad + \beta_5 \times (p \cdot \theta) \\
  & \quad + \boldsymbol{\gamma} \mathbf{FE} + \varepsilon,
\end{split}
\end{equation}

\begin{equation}
\begin{split}
\text{Model 3:}\quad 
Y &= \beta_0 + \beta_1 \times p + \beta_2 \times \beta \\
  & \quad + \beta_3 \times \theta + \beta_4 \times k \\
  & \quad + \beta_5 \times (p \cdot \theta) \\
  & \quad + \beta_6 \times (\beta \cdot k) \\
  & \quad + \beta_7 \times (\beta \cdot p) \\
  & \quad + \beta_8 \times (\beta \cdot \theta) \\
  & \quad + \boldsymbol{\gamma} \mathbf{FE} + \varepsilon,
\end{split}
\end{equation}

\begin{equation}
\begin{split}
\text{Model 4:}\quad 
Y &= \beta_0 + \beta_1 \times p + \beta_2 \times \beta \\
  & \quad + \beta_3 \times \theta + \beta_4 \times k \\
  & \quad + \beta_5 \times (p \cdot \theta) \\
  & \quad + \beta_6 \times (\beta \cdot k) \\
  & \quad + \beta_9 \times (p \cdot k) \\
  & \quad + \beta_{10} \times (\theta \cdot k) \\
  & \quad + \boldsymbol{\gamma} \mathbf{FE} + \varepsilon.
\end{split}
\end{equation}

\begin{equation}
\begin{split}
\text{Model 5:}\quad 
Y &= \beta_0 + \beta_1 \times p + \beta_2 \times \beta \\
  & \quad + \beta_3 \times \theta + \beta_4 \times k \\
  & \quad + \beta_5 \times (p \cdot \theta) \\
  & \quad + \beta_6 \times (\beta \cdot k) \\
  & \quad + \beta_7 \times (\beta \cdot p) \\
  & \quad + \beta_8 \times (\beta \cdot \theta) \\
  & \quad + \beta_9 \times (p \cdot k) \\
  & \quad + \beta_{10} \times (\theta \cdot k) \\
  & \quad + \boldsymbol{\gamma} \mathbf{FE} + \varepsilon.
\end{split}
\end{equation}

To systematically investigate the drivers of bias, we estimate six nested OLS models. Model~0 serves as a specialized baseline, analyzing only the static environment with no historical data ($N=600$ observations). This model allows us to isolate the fundamental $\beta$ effects without the confounding influence of path complexity. Model~1 includes the main effects of $p$, $\beta$, $\theta$, and $k$, together with fixed effects for 4 price trajectories. Model~2 augments this baseline with an interaction between $p$ and $\theta$, capturing how sensitivity to price depends on agent type. Model~3 introduces interactions between $\beta$ and the other regressors, allowing us to test whether the presence of stronger interdependence intensity systematically amplifies or dampens deviations. Model~4 replaces these with interactions between $k$ and the other regressors, to assess whether a longer context window reshapes the influence of price and type on expectations. Finally, Model~5 includes all two-way interaction terms simultaneously to assess the robustness of the individual interaction effects. To sum up, these six models provide a comprehensive view of how internal heterogeneity and external conditions shape agents’ deviations from REE.

\begin{table*}[!htb]
\centering
\caption{Regression Analysis with Price-Path Fixed Effects}
\label{table:full_sample_analysis}
\begin{tabular}{l c c c c c c}
\toprule
Variable & Model 0 & Model 1 & Model 2 & Model 3 & Model 4 & Model 5 \\
\midrule
$p$          & 19.58*** & 17.31*** & 20.11*** & 12.96*** & 27.38*** & 20.23*** \\
             & (1.15)   & (0.31)   & (0.62)   & (0.62)   & (0.71)   & (0.71)   \\[2pt]
$\beta$      & -0.21    & 0.32     & 0.32     & -5.84*** & 0.41     & -5.84*** \\
             & (0.70)   & (0.18)   & (0.18)   & (0.55)   & (0.31)   & (0.52)   \\[2pt]
$\theta$     & 6.29***  & 3.67***  & 6.52***  & 7.45***  & 9.38***  & 10.30*** \\
             & (1.22)   & (0.31)   & (0.62)   & (0.67)   & (0.73)   & (0.77)   \\[2pt]
$k$          & --       & 2.06***  & 2.06***  & 2.15***  & 13.06*** & 13.06*** \\
             &          & (0.25)   & (0.25)   & (0.30)   & (0.62)   & (0.60)   \\[2pt]
$p \times \theta$ & --  & --       & -5.70*** & -5.70*** & -5.70*** & -5.70*** \\
             &          &          & (1.07)   & (1.03)   & (1.01)   & (0.97)   \\[2pt]
$\beta \times p$  & --  & --       & --       & 14.29*** & --       & 14.29*** \\
             &          &          &          & (0.59)   &          & (0.55)   \\[2pt]
$\beta \times \theta$ & -- & --    & --       & -1.84*** & --       & -1.84*** \\
             &          &          &          & (0.60)   &          & (0.57)   \\[2pt]
$\beta \times k$  & --  & --       & --       & -0.18    & -0.18    & -0.18    \\
             &          &          &          & (0.43)   & (0.43)   & (0.41)   \\[2pt]
$p \times k$      & --  & --       & --       & --       & -15.76***& -15.76***\\
             &          &          &          &          & (0.73)   & (0.68)   \\[2pt]
$\theta \times k$ & --  & --       & --       & --       & -6.19*** & -6.19*** \\
             &          &          &          &          & (0.73)   & (0.71)   \\
\midrule
Price-path FE & Inc.    & Inc.     & Inc.     & Inc.     & Inc.     & Inc.     \\
\# Obs       & 600      & 7,800    & 7,800    & 7,800    & 7,800    & 7,800    \\
$R^2$        & 0.401    & 0.385    & 0.388    & 0.447    & 0.442    & 0.501    \\
\bottomrule
\multicolumn{7}{l}{\footnotesize Note: *** $p<0.01$; ** $p<0.05$. HC3 robust SEs in parentheses.} \\
\end{tabular}

\end{table*}

The regression results are reported in Table~\ref{table:full_sample_analysis}. The standardized regression analysis clarifies that the main driver of behavior is the sequential structure of the context, while interdependence strength acts as an amplifier rather than an independent cause.
In our full sample with price path fixed effects, the main effect of $\beta$ is small and statistically indistinguishable from zero in the baseline specifications. However, $\beta$ shows strong and significant interactions with the price level and with the trend signal, indicating that higher strategic interdependence magnifies agents’ sensitivity to the same sequential patterns.
The influence of the trend structure itself remains robust and substantial after controlling for all $\beta$ interactions.
The interaction terms between context length and price, as well as context length and the trend signal, are both highly significant and have meaningful magnitudes. This pattern explains why the jump sequences in Table~\ref{table:metrics_gpt_5} sometimes show improvement at low $\beta$ but fail at high $\beta$. The sequential structure of the history is the primary force pulling behavior away from equilibrium, and higher $\beta$ simply intensifies this effect rather than replacing it.

\section{Finite State Machine}\label{appendix:finite_state_machine}

We detail the experimental workflow in this section.

\subsection{Experimental Workflow Formalization}

As shown in Figure \ref{figure:finite_state_machine}, we defined the core components of our FSM-based workflow to formalize the experimental protocol into a verifiable computational model. The FSM framework is a well-established model in computer science and systems engineering for representing discrete event systems. It is formally defined by the quadruple $\mathcal{M} = (S, E, \delta, S_0)$, where:

\begin{itemize}
    \item $S$ is a finite set of states, representing the instantaneous configuration or operational phase of the system at any given moment. Each state encapsulates all critical information and operational permissions, ensuring logical isolation between states.
    \item $E$ is a finite set of events, serving as atomic signals that trigger state transitions. An event is instantaneous and non-durable, marking the fulfillment of a specific condition or the completion of an operation.
    \item $\delta:S \times E \rightarrow S$ is the state transition function, which precisely defines how the system transitions from the current state $s \in S$ to the next state $s' \in S$ upon receiving an event $e \in E$. This function is the core computational rule governing the model's evolution.
    \item $S_0$ is a unique initial state, representing the deterministic starting point of the entire experimental protocol. This ensures the predictability and consistent execution of the experiment across all runs.
\end{itemize}

Our workflow is structured around a sequence of well-defined states, event triggers, and state transitions, which are detailed below.

\subsection{States}

A state is the instantaneous configuration of the FSM at any moment, representing a specific operational stage of the experiment. We define the state set $S = \{S_0, S_1, S_2, S_3, S_4\}$, where each state corresponds to a unique and logically exclusive phase:

\begin{itemize}
    \item $S_0$: \textbf{Initialization}. It is responsible for setting all key parameters, including global environment parameters (e.g., interdependence strength $\beta$, payoff function $u_i(\cdot)$ (Equation~\ref{equation:payoff})) and private parameters for each agent (e.g., private value $\theta_i$). 
    \item $S_1$: \textbf{Information Broadcast}. This state defines the unidirectional information synchronization phase from the environment to the agents. The environment acts as a central coordinator, distributing public information for the current round to all agents. This information typically includes the current public cost $p_t$ and public feedback from the previous round (e.g., the total number of participants $N_{t-1}$). This state establishes the information baseline for a new round of the game.
    \item $S_2$: \textbf{Agent Decision-Making}. This state represents a distributed and parallel decision process. Each independent agent $Agent_i$, given its private information $\theta_i$, currently observable public cost $p_t$, and historical data, independently executes its internal decision strategy. We further decompose the decision process into two subprocesses: participant expectation and decision making. All agent's decision logic from a simple binary choice to a complex process based on quantitative analysis and reasoning. Crucially, all agents' decisions are chronologically synchronous and informationally independent; they occur simultaneously and independently, without access to real-time decisions of other agents.
    \item $S_3$: \textbf{Outcome Aggregation \& Payoff}. This state defines the quantification of collective behavior and the calculation of individual rewards. In this state, the system collects all agents' actions $\{a_t\}$, aggregates them to form the collective outcome for the current round (i.e., total participants $N_t = \sum_i a_t$), and calculates each agent's payoff $u_j$ based on the payoff function defined in Equation~\ref{equation:payoff}. This payoff can be considered a reward signal for evaluating agent strategies or for future learning.
    \item $S_4$: \textbf{Termination}. This is a final absorbing state, which marks the satisfaction of all experimental conditions. The system will not perform any further transitions from this state, ensuring the deterministic conclusion of the experiment.
\end{itemize}

\subsection{Events}

Events are atomic signals that trigger state transitions. They represent instantaneous, non-durable actions or outcomes that transition the system from one state to another. These events are generated by external entities, such as the environment or the agent population.

\begin{itemize}
    \item $E_0$: \textbf{Experiment Start}, triggered by the environment, marking the logical end of the initialization phase.
    \item $E_1$: \textbf{Broadcast Complete}, triggered by the environment, indicating the completion of information synchronization.
    \item $E_2$: \textbf{All Decisions Complete}, a joint and synchronous event triggered by the agent population, marking the collective synchronization of the parallel decision process. This is a critical synchronization point for the system's transition from a distributed state to a centralized one.
    \item $E_3$: \textbf{Results Calculated}, triggered by the environment, marking the deterministic generation of collective results and individual payoffs for the current round.
    \item $E_4$: \textbf{Termination Met}, triggered by the environment, indicating that the experiment's termination conditions have been met.
\end{itemize}

\subsection{State Transition Function}

We use the state transition function $\delta$ to define how the system transitions from one state to the next based on the received event. The formal representation is $\delta:S \times E \rightarrow S$. This function formalizes the experimental protocol into a verifiable computational model, with the sequence as follows:

\begin{itemize}
    \item $\delta(S_0, E_0) \rightarrow S_1$: After initialization is complete, the experiment enters the public cost publication state.
    \item $\delta(S_1, E_1) \rightarrow S_2$: After the public information broadcast is complete, the system enters the parallel agent decision-making state.
    \item $\delta(S_2, E_2) \rightarrow S_3$: After all agents have made their decisions, the system enters the results calculation state.
\end{itemize}

In state $S_3$, the transition path is determined by the experiment's termination conditions:

\begin{itemize}
    \item $\delta(S_3, E_3) \rightarrow S_1$: If the experiment has not yet ended, the workflow returns to $S_1$ to begin a new round.
    \item $\delta(S_3, E_4) \rightarrow S_4$: If the experiment's termination conditions are met, the workflow enters a final absorbing state, and the experiment concludes.
\end{itemize}

\subsection{Workflow Execution}

The experiment began with an initial parameter setup phase, followed by a cyclical iterative process until the termination conditions are met. In this manner, we formalized the experimental protocol into a verifiable computational model, clearly defining the interfaces and interaction logic.

\begin{itemize}
    \item \textbf{Initial Phase}: The workflow is in state $S_0$. In this state, the environment specifies the interdependence strength ($\beta$), public cost sequence ($P$), and payoff function (Equation~\ref{equation:payoff}). Concurrently, each agent is assigned a private value ($\theta_i$). When all parameters are set, event $E_0$ is triggered, and the state transitions to $S_1$.
    \item \textbf{Public Cost Publication Phase}: The workflow enters state $S_1$. The environment selects the public cost for the current round from the preset cost sequence and broadcasts it to all agents. In this phase, the environment also distributes feedback from the previous round (if available). When the public cost broadcast is complete, event $E_1$ is triggered, and the state transitions to $S_2$.
    \item \textbf{Agent Decision-Making Phase}: The workflow enters state $S_2$. All agents analyze and reason based on their limited information (including historical public costs, their private values, and past payoff records) to form an expectation of the number of participants. In this phase, each agent is independent and cannot access information about other agents. Only when all agents have completed their decisions is event $E_2$ triggered, and the state transitions to $S_3$.
    \item \textbf{Results Calculation Phase}: The workflow enters state $S_3$. The environment collects all agents' decisions, calculates the actual number of participants for the current round, and, based on this, computes each agent's actual payoff. When the calculation is complete, event $E_3$ is triggered.
    \item \textbf{Looping and Termination}: The workflow starting a new cycle. If there are no more public costs, event $E_4$ is triggered, and the workflow finally transitions to a termination state, ending the experiment.
\end{itemize}

\section{Price Sequences}\label{appendix:price_sequences}

\subsection{Equilibrium Solution of the Theoretical Model}

The price equilibrium under the REE theory is solved through an iterative process. Under the assumption that all agents are rational, they will attempt to predict the expectations and behavior of other agents. In this framework, the expectations of all agents and the resulting equilibrium state are unique.

The decision-making process for any given agent (regardless of their individual private value $\theta$) can be reasoned starting from the agents with the most extreme private value. In the scenario defined in this paper, this reasoning begins with the highest price. As long as an agent's utility is positive, they will choose to participate. As the number of participants increases, the utility of all agents in the network will improve. Subsequently, the system calculates the utility for the next agent with an extreme private value. This process continues to iterate until the utility of all agents has been evaluated.

Based on this method, we iteratively solved for the REE theoretical solution using the parameter $\beta$ and the private value $\theta_i$ for each agent $i$. It is worth noting that to avoid boundary conditions in the calculation. Specifically, cases where the utility is exactly zero, leading to an ambiguous preference for an agent to participate or not. We subtracted 0.01 from the final utility value. This ensures that the utility value is always strictly positive or negative, making the agent's decision unambiguous.

\subsection{Generation of Experimental Price Settings}

We computationally derived the prices corresponding to a specific equilibrium number of participants for two different values of $\beta$. For simplicity of presentation, we only list the equilibrium solutions at intervals of 10 participants.

\begin{itemize}
    \item When $\beta = 0.25$, the prices under the REE condition to achieve an equilibrium number of participants of 0, 10, 20, 30, 40, and 50 are 49.99, 42.49, 34.99, 27.49, 19.99, and 12.49, respectively.
    \item When $\beta = 0.75$, the prices under the REE condition to achieve an equilibrium number of participants of 0, 10, 20, 30, 40, and 50 are 49.99, 47.49, 44.99, 42.49, 39.99, and 37.49, respectively.
\end{itemize}

\subsection{Experimental Scenario Design and Price Sequences}

The experiment manipulates the temporal ordering of the same set of prices to form two types of sequences.

\paragraph{Monotonic Sequences}

Prices change in a single direction, exhibiting a clear local trend.

\begin{itemize}
    \item \textbf{Decreasing Sequence}: Prices are arranged from the highest equilibrium number of participants to the lowest (indices: 0, 1, 2, 3, 4, 5).
    \item \textbf{Increasing Sequence}: Prices are arranged from the lowest equilibrium number of participants to the highest (indices: 5, 4, 3, 2, 1, 0).
\end{itemize}

\paragraph{Jump Sequences}

Adjacent prices frequently switch between high and low values, forming no stable trend.

\begin{itemize}
    \item \textbf{Converging Sequence}: Prices converge from the two ends to the middle (indices: 5, 0, 4, 1, 3, 2).
    \item \textbf{Diverging Sequence}: Prices diverge from the middle to the two ends (indices: 2, 3, 1, 4, 0, 5).
\end{itemize}

\paragraph{Specific Price Sequences}

\begin{itemize}
    \item When $\beta=0.25$:
    \begin{itemize}
        \item Decreasing sequence: [49.99, 42.49, 34.99, 27.49, 19.99, 12.49]
        \item Increasing sequence: [12.49, 19.99, 27.49, 34.99, 42.49, 49.99]
        \item Converging sequence: [49.99, 12.49, 42.49, 19.99, 34.99, 27.49]
        \item Diverging sequence: [27.49, 34.99, 19.99, 42.49, 12.49, 49.99]
    \end{itemize}
    \item When $\beta=0.75$:
    \begin{itemize}
        \item Decreasing sequence: [49.99, 47.49, 44.99, 42.49, 39.99, 37.49]
        \item Increasing sequence: [37.49, 39.99, 42.49, 44.99, 47.49, 49.99]
        \item Converging sequence: [49.99, 37.49, 47.49, 39.99, 44.99, 42.49]
        \item Diverging sequence: [42.49, 44.99, 39.99, 47.49, 37.49, 49.99]
    \end{itemize}
\end{itemize}

\section{Experimental Setup}\label{appendix:experimental_setup}

To ensure consistent and high-quality model outputs, we adhered to the official recommendations for each LLM's decoding parameters. The specific settings for each model are summarized below:

\begin{itemize}
    \item \textbf{GPT-5}: temperature=0.7, top\_p=1.0, top\_k=20, presence\_penalty=0
    \item \textbf{Qwen3-Plus}: temperature=0.7, top\_p=0.8, top\_k=20, presence\_penalty=1.5
\end{itemize}

These parameters were selected to optimize the models' performance and provide a stable baseline for our research, following the guidelines provided in their respective official documentation.

To investigate the impact of the decoding strategy on the agent's decision-making behavior, we conducted a supplementary analysis by varying the temperature parameter. We re-ran all experiments with the temperature for each model uniformly set to 0.35, while keeping all other parameters constant. For the results, please refer to Appendix~\ref{appendix:results}. The choice of 0.35 was deliberate, as it introduces a small degree of randomness necessary to prevent the models from falling into deterministic output loops in complex scenarios.

\section{Additional Experimental Results}\label{appendix:results}

\subsection{Performance across Different Model}

To systematically test the transferability and robustness of our main empirical conclusions to differences in the capabilities of the base LLM, this section replaces the GPT-5 model used in the main text with the Qwen3-Plus model and replicates all experimental scenarios.

\begin{table}[htb]
\centering
\small
\caption{Qwen3-Plus: Equilibrium deviation and equilibrium determination coefficient across all trajectories.}
\label{table:metrics_qwen_plus}
\begin{tabular}{cc ccc}
\toprule
$\bm{\mathbf{\beta}}$ & \textbf{Trajectory} & $\bm{\mathbf{k}}$ & $\bm{\mathrm{ED}}$ & $\bm{\mathrm{R}^2_{eq}}$ \\
\midrule
\multirow{13}{*}{0.25} & Static & -- & 13.275 & 0.396 \\
\cmidrule{2-5}
 & \multirow{3}{*}{Decreasing} & 0 & 10.000 & 0.657 \\
 &                             & 3 & 10.983 & 0.586 \\
 &                             & 6 & 9.157 & 0.713 \\
\cmidrule{2-5}
 & \multirow{3}{*}{Increasing} & 0 & 14.794 & 0.250 \\
 &                             & 3 & 9.178 & 0.711 \\
 &                             & 6 & 10.351 & 0.633 \\
\cmidrule{2-5}
 & \multirow{3}{*}{Converging} & 0 & 15.974 & 0.125 \\
 &                             & 3 & 14.674 & 0.262 \\
 &                             & 6 & 14.363 & 0.293 \\
\cmidrule{2-5}
 & \multirow{3}{*}{Diverging}  & 0 & 16.321 & 0.087 \\
 &                             & 3 & 11.876 & 0.516 \\
 &                             & 6 & 10.604 & 0.615 \\
\midrule
\multirow{13}{*}{0.75} & Static & -- & 17.364 & -0.034 \\
\cmidrule{2-5}
 & \multirow{3}{*}{Decreasing} & 0 & 12.374 & 0.475 \\
 &                             & 3 & 12.812 & 0.437 \\
 &                             & 6 & 10.245 & 0.640 \\
\cmidrule{2-5}
 & \multirow{3}{*}{Increasing} & 0 & 18.225 & -0.139 \\
 &                             & 3 & 13.677 & 0.359 \\
 &                             & 6 & 13.544 & 0.371 \\
\cmidrule{2-5}
 & \multirow{3}{*}{Converging} & 0 & 17.281 & -0.024 \\
 &                             & 3 & 15.118 & 0.216 \\
 &                             & 6 & 15.344 & 0.193 \\
\cmidrule{2-5}
 & \multirow{3}{*}{Diverging}  & 0 & 20.132 & -0.390 \\
 &                             & 3 & 16.758 & 0.037 \\
 &                             & 6 & 15.715 & 0.153 \\
\bottomrule
\end{tabular}
\end{table}

\begin{table*}[htb]
\centering
\small
\caption{Qwen3-Plus: OLS regression of observed participation on mapped price.}
\label{tab:regression_summary_qwen_plus}
\begin{tabular}{c cc ccc}
\toprule
$\bm{\mathbf{\beta}}$ & \textbf{Trajectory} & $\bm{\mathbf{k}}$ & $\bm{\alpha}$ (t) & $\bm{\gamma}$ (t) & $\bm{\mathbf{R^2 / \bar{R}^2}}$ \\
\midrule
\multirow{13}{*}{0.25} & Static & -- & 36.19$^{***}$ (33.08) & -6.71$^{***}$ (-18.58) & 0.537 / 0.535 \\
\cmidrule{2-6}
 & \multirow{3}{*}{Decreasing} & 0 & 41.02$^{***}$ (46.35) & -7.66$^{***}$ (-26.19) & 0.697 / 0.696 \\
 & & 3 & 46.05$^{***}$ (42.25) & -8.35$^{***}$ (-23.19) & 0.643 / 0.642 \\
 & & 6 & 48.36$^{***}$ (53.76) & -8.73$^{***}$ (-29.37) & 0.743 / 0.742 \\
\cmidrule{2-6}
 & \multirow{3}{*}{Increasing} & 0 & 38.47$^{***}$ (30.42) & -5.21$^{***}$ (-12.46) & 0.343 / 0.340 \\
 & & 3 & 40.48$^{***}$ (53.01) & -7.52$^{***}$ (-29.78) & 0.749 / 0.748 \\
 & & 6 & 38.28$^{***}$ (46.83) & -6.90$^{***}$ (-25.57) & 0.687 / 0.686 \\
\cmidrule{2-6}
 & \multirow{3}{*}{Converging} & 0 & 36.72$^{***}$ (27.20) & -4.68$^{***}$ (-10.51) & 0.270 / 0.268 \\
 & & 3 & 32.64$^{***}$ (29.61) & -4.59$^{***}$ (-12.60) & 0.348 / 0.345 \\
 & & 6 & 32.66$^{***}$ (31.07) & -4.47$^{***}$ (-12.88) & 0.358 / 0.355 \\
\cmidrule{2-6}
 & \multirow{3}{*}{Diverging} & 0 & 35.70$^{***}$ (26.61) & -4.27$^{***}$ (-9.64) & 0.238 / 0.235 \\
 & & 3 & 37.81$^{***}$ (38.36) & -6.25$^{***}$ (-19.20) & 0.553 / 0.552 \\
 & & 6 & 39.78$^{***}$ (43.67) & -6.97$^{***}$ (-23.17) & 0.643 / 0.642 \\
\midrule
\multirow{13}{*}{0.75} & Static & -- & 28.77$^{***}$ (22.78) & -3.35$^{***}$ (-8.02) & 0.178 / 0.175 \\
\cmidrule{2-6}
 & \multirow{3}{*}{Decreasing} & 0 & 37.45$^{***}$ (39.17) & -5.24$^{***}$ (-16.60) & 0.481 / 0.479 \\
 & & 3 & 46.22$^{***}$ (43.56) & -6.57$^{***}$ (-18.74) & 0.541 / 0.539 \\
 & & 6 & 45.92$^{***}$ (54.34) & -7.02$^{***}$ (-25.15) & 0.680 / 0.679 \\
\cmidrule{2-6}
 & \multirow{3}{*}{Increasing} & 0 & 39.01$^{***}$ (34.55) & -2.66$^{***}$ (-7.14) & 0.146 / 0.143 \\
 & & 3 & 37.02$^{***}$ (37.03) & -4.41$^{***}$ (-13.34) & 0.374 / 0.372 \\
 & & 6 & 38.31$^{***}$ (39.58) & -4.45$^{***}$ (-13.92) & 0.394 / 0.392 \\
\cmidrule{2-6}
 & \multirow{3}{*}{Converging} & 0 & 36.93$^{***}$ (30.49) & -3.05$^{***}$ (-7.62) & 0.163 / 0.160 \\
 & & 3 & 32.75$^{***}$ (36.38) & -2.80$^{***}$ (-9.42) & 0.229 / 0.227 \\
 & & 6 & 33.11$^{***}$ (32.36) & -3.17$^{***}$ (-9.37) & 0.228 / 0.225 \\
\cmidrule{2-6}
 & \multirow{3}{*}{Diverging} & 0 & 32.12$^{***}$ (26.96) & -0.83$^{*}$ (-2.11) & 0.015 / 0.011 \\
 & & 3 & 32.57$^{***}$ (28.32) & -2.71$^{***}$ (-7.15) & 0.146 / 0.143 \\
 & & 6 & 34.84$^{***}$ (32.65) & -3.18$^{***}$ (-9.37) & 0.215 / 0.212 \\
\bottomrule
\multicolumn{6}{l}{\footnotesize Note: $t$-statistics in parentheses. ${}^{***}p<0.001$, ${}^{**}p<0.01$, ${}^{*}p<0.05$.} \\
\end{tabular}
\end{table*}

\begin{figure*}[htb]
    \centering
    \includegraphics[width=1.0\linewidth]{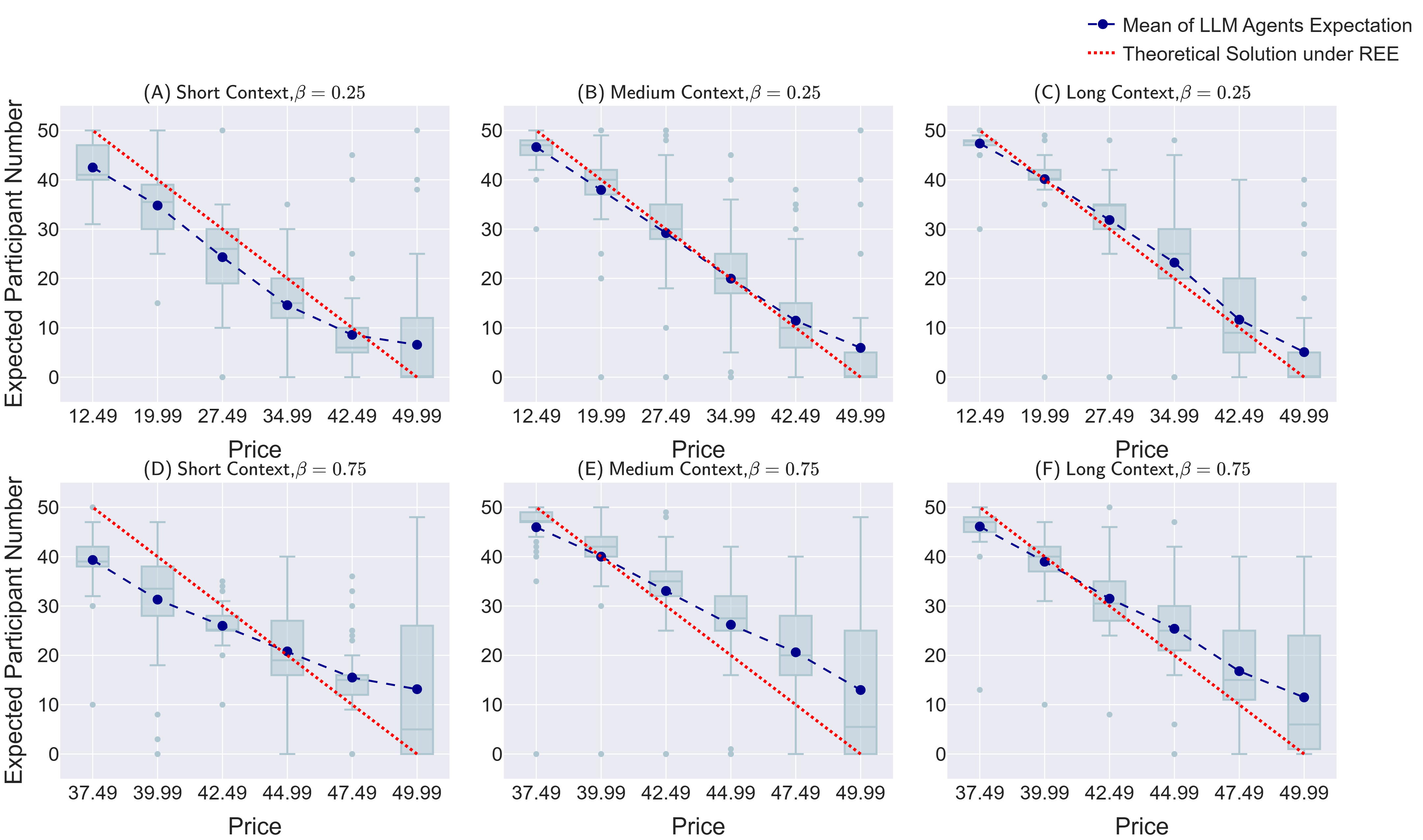}
    \caption{Qwen3-Plus: Monotonic decreasing price trajectory.}
    \label{fig:results_qwen_plus_decreasing}
\end{figure*}

\begin{figure*}[htb]
    \centering
    \includegraphics[width=1.0\linewidth]{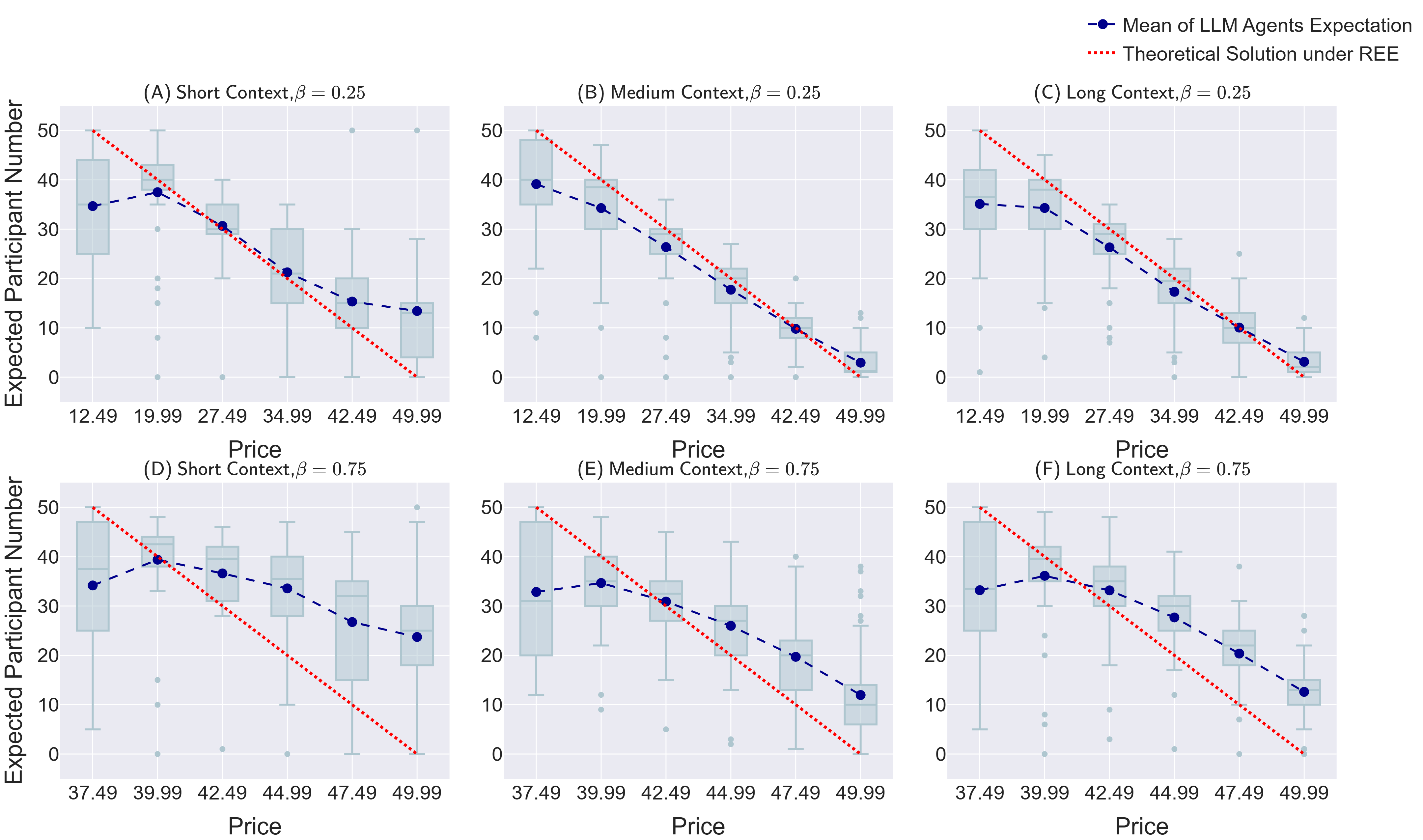}
    \caption{Qwen3-Plus: Monotonic increasing price trajectory.}
    \label{fig:results_qwen_plus_increasing}
\end{figure*}

\begin{figure*}[htb]
    \centering
    \includegraphics[width=1.0\linewidth]{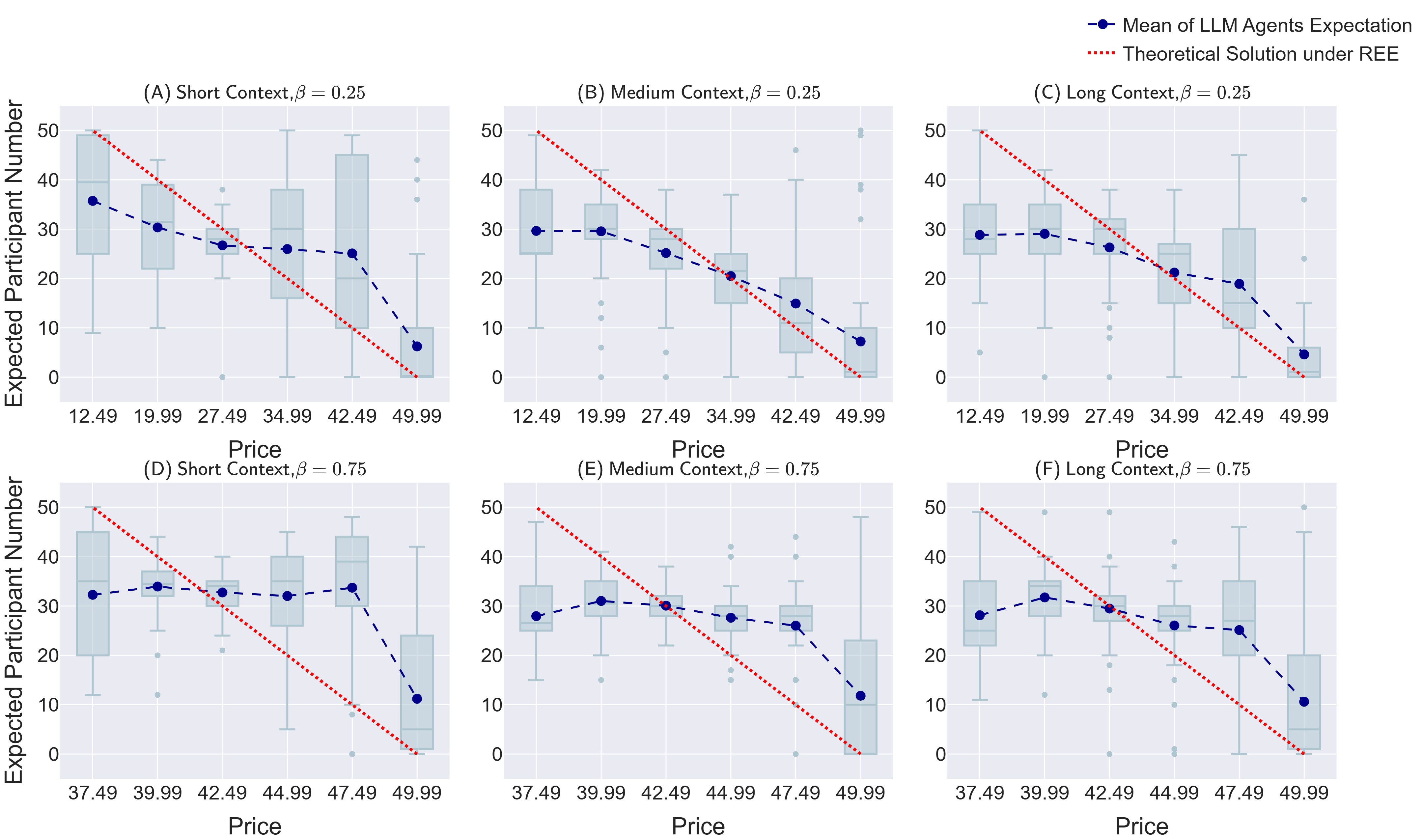}
    \caption{Qwen3-Plus: Jump converging price trajectory.}
    \label{fig:results_qwen_plus_converging}
\end{figure*}

\begin{figure*}[htb]
    \centering
    \includegraphics[width=1.0\linewidth]{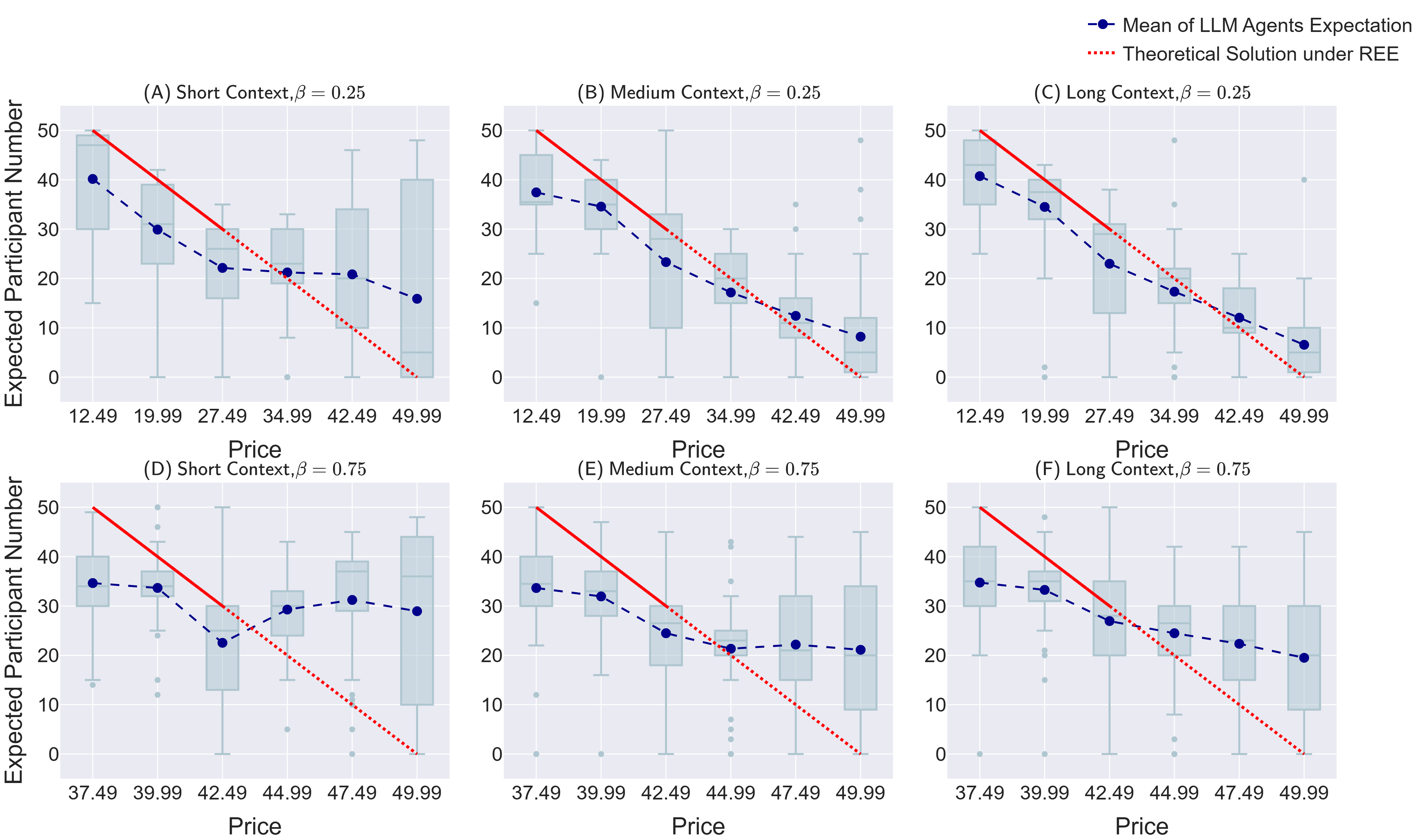}
    \caption{Qwen3-Plus: Jump diverging price trajectory.}
    \label{fig:results_qwen_plus_diverging}
\end{figure*}

Figures \ref{fig:results_qwen_plus_decreasing}, \ref{fig:results_qwen_plus_increasing}, \ref{fig:results_qwen_plus_converging} and \ref{fig:results_qwen_plus_diverging} display the decision-making results based on the Qwen3-Plus agent. A qualitative analysis reveals that the Qwen3-Plus agent's response patterns to network effect strength, historical state window length, and price sequences show high consistency in qualitative trends with the GPT-5 agent used in the main text. Specifically, agents from both models exhibit superior and more concentrated decision-making behavior in weak network effect environments, whereas their behavior is worse and more divergent in strong network effect environments.

However, there are significant systematic differences at the quantitative level. Compared to the GPT-5 agent, the Qwen3-Plus agent is generally less sensitive to changes in price trends, which leads to a more pronounced deviation of its group's expected mean from the theoretical equilibrium solution. Furthermore, the expected dispersion of the Qwen3-Plus agent group is systematically higher. This phenomenon may inherently reflect the Qwen3-Plus model's relative deficiency in processing complex contextual information and reasoning, which affects the consistency and accuracy of its decisions.

\subsection{Comprehensive Robustness Results}\label{appendix:robustness_check}

To evaluate the sensitivity of our main conclusions to stochastic perturbations in the model's decoding strategy, this section conducts a sensitivity analysis on the core experimental results of the GPT-5 model. We adjust the decoding temperature parameter (T) from the benchmark T=0.7 in the main text to T=0.35 to significantly reduce randomness during the sampling process. All other parameter configurations are maintained consistently with the optimal parameter set described in Appendix \ref{appendix:experimental_setup}.

\begin{figure*}[!htb]
    \centering
    \includegraphics[width=1.0\linewidth]{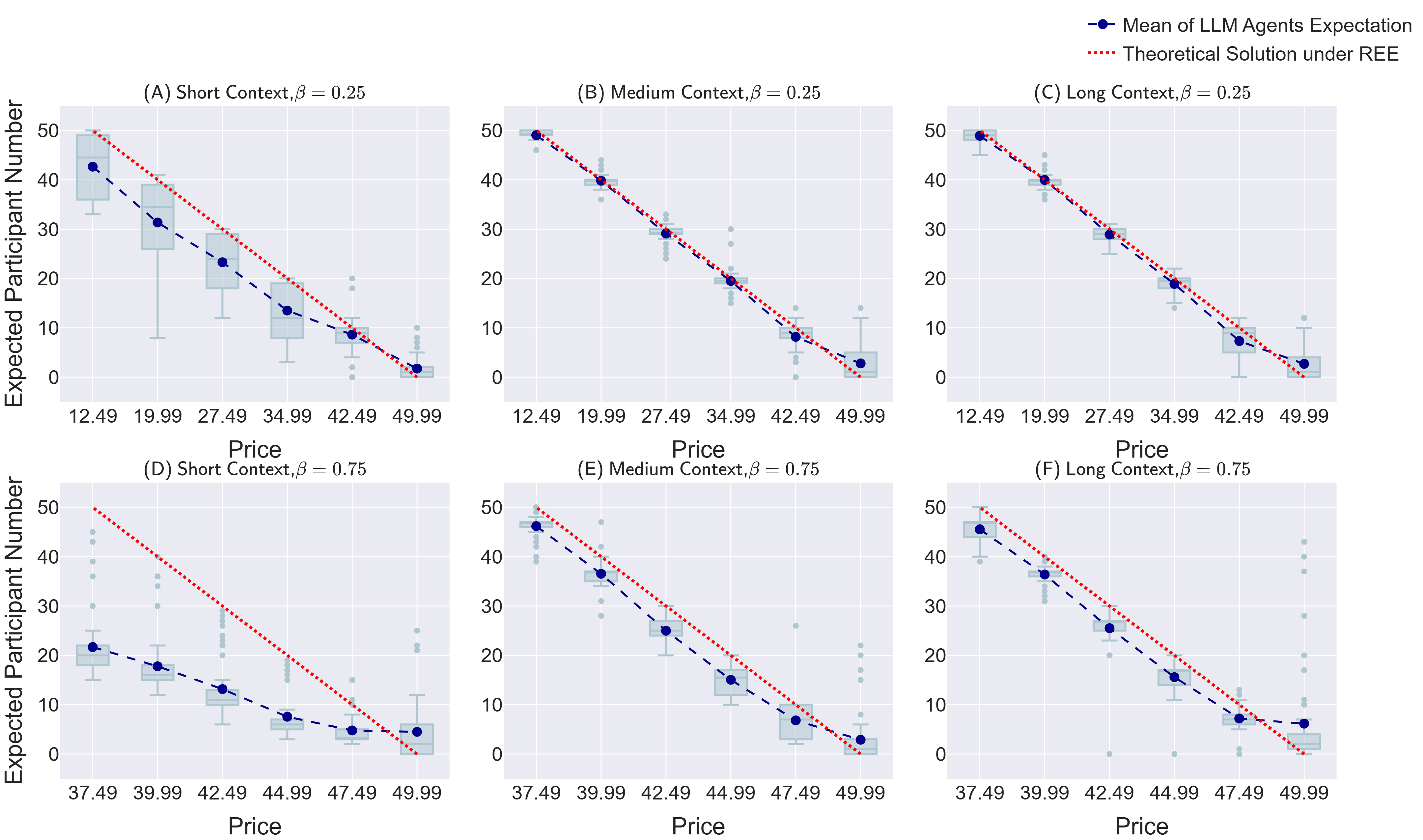}
    \caption{GPT-5: Monotonic decreasing price trajectory. Temperature=0.35.}
    \label{fig:robustness_results_decreaing_gpt_5}
\end{figure*}

\begin{figure*}[!htb]
    \centering
    \includegraphics[width=1.0\linewidth]{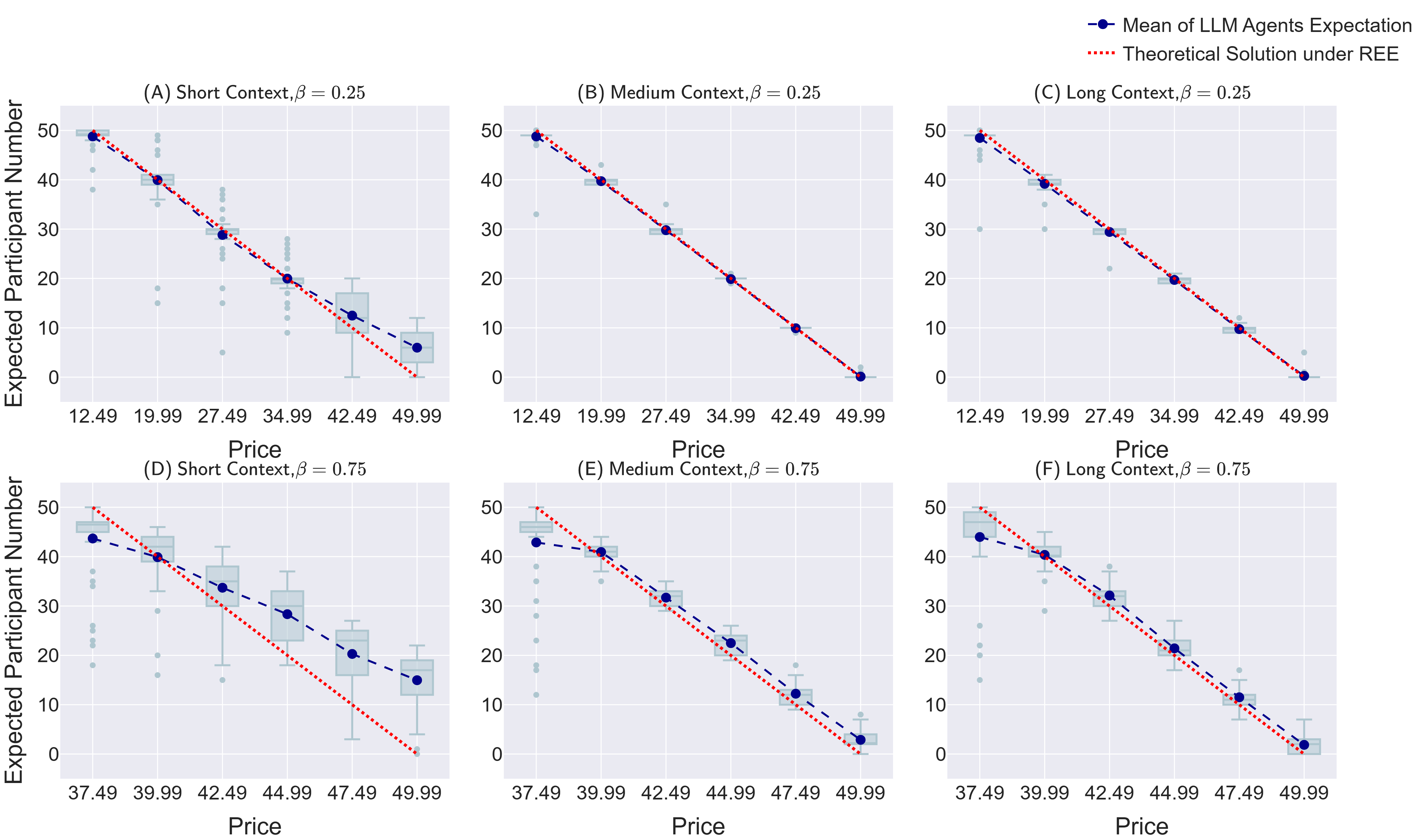}
    \caption{GPT-5: Monotonic increasing price trajectory. Temperature=0.35.}
    \label{fig:robustness_results_increaing_gpt_5}
\end{figure*}

\begin{figure*}[!htb]
    \centering
    \includegraphics[width=1.0\linewidth]{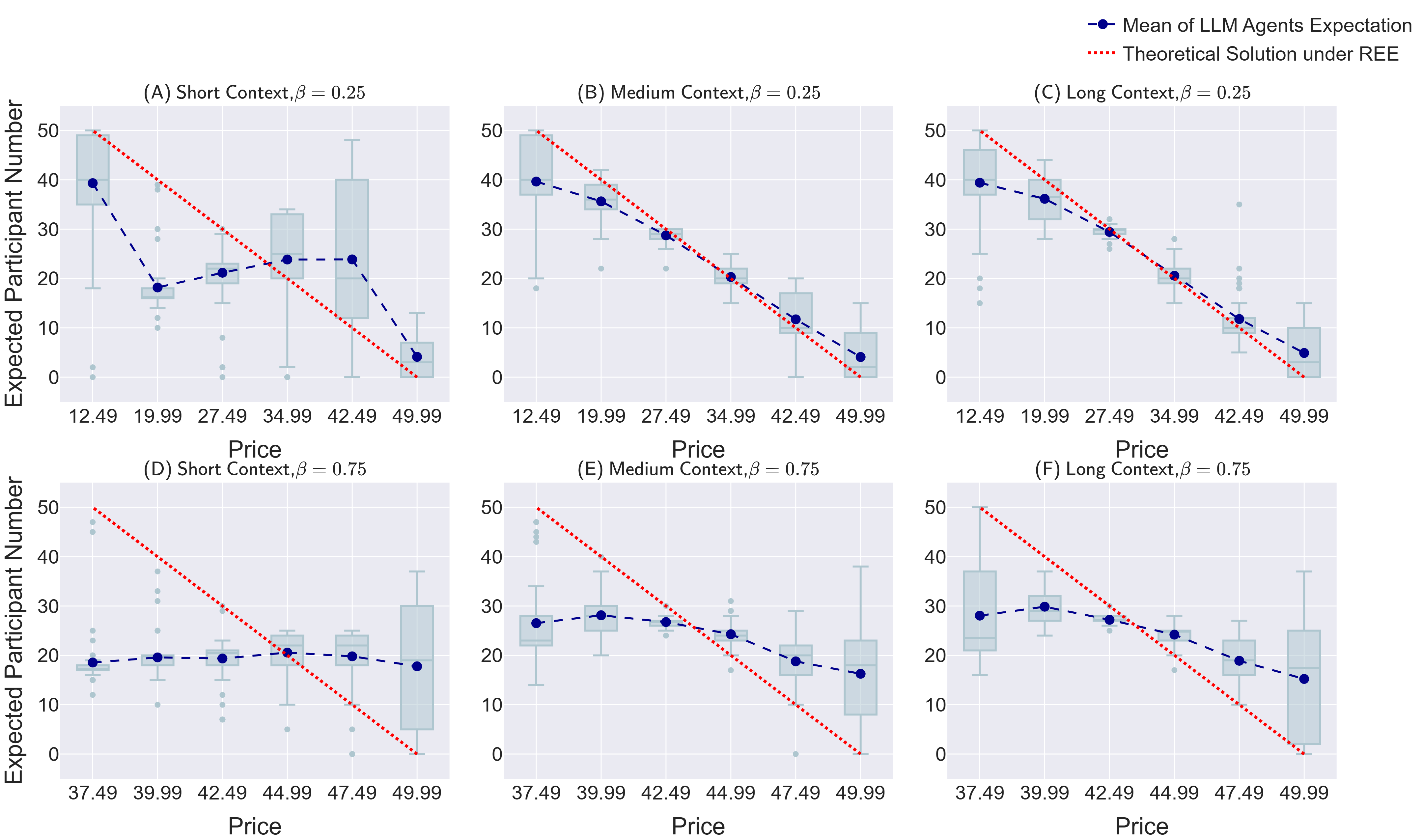}
    \caption{GPT-5: Jump converging price trajectory. Temperature=0.35.}
    \label{fig:robustness_results_converging_gpt_5}
\end{figure*}

\begin{figure*}[!htb]
    \centering
    \includegraphics[width=1.0\linewidth]{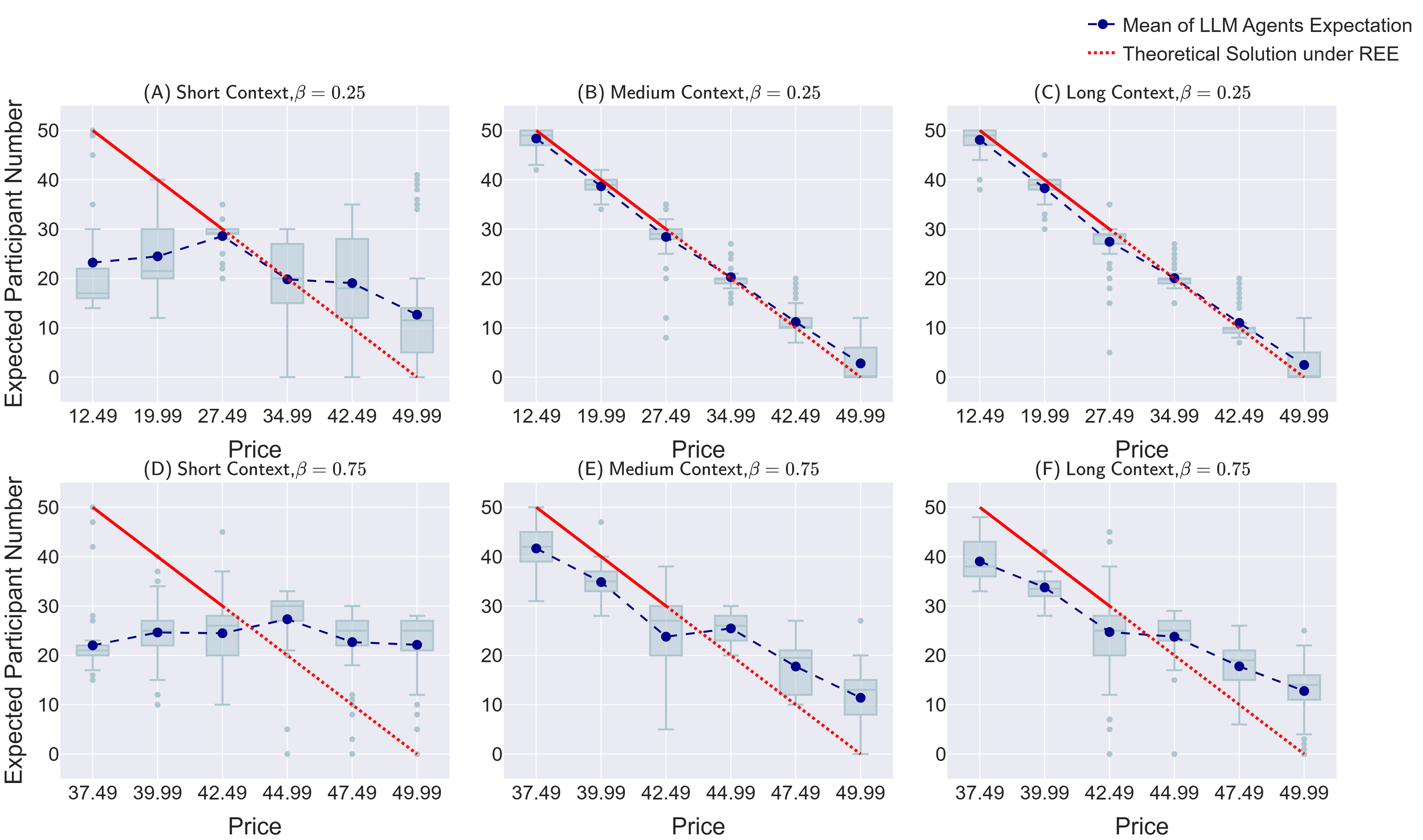}
    \caption{GPT-5: Jump diverging price trajectory. Temperature=0.35.}
    \label{fig:robustness_results_diverging_gpt_5}
\end{figure*}

Figures \ref{fig:robustness_results_decreaing_gpt_5}, \ref{fig:robustness_results_increaing_gpt_5}, \ref{fig:robustness_results_converging_gpt_5} and \ref{fig:robustness_results_diverging_gpt_5} show the experimental results under the condition of temperature=0.35. The analysis indicates that a reduction in decoding temperature does not fundamentally alter the agent's core decision-making behavior patterns. The qualitative responses of the agent group to different network effect strengths and historical window lengths remain highly consistent with the benchmark results at temperature=0.7.

\end{document}